\documentclass[runningheads]{llncs}

\usepackage{eccv}

\usepackage{eccvabbrv}

\usepackage{graphicx}
\usepackage{booktabs}
\usepackage{comment}
\usepackage{makecell}
\usepackage{colortbl}
\usepackage{color, xcolor}
\usepackage{multirow} 
\usepackage{float} 
\usepackage{bbding}
\usepackage[accsupp]{axessibility}  

\usepackage{hyperref}

\usepackage{orcidlink}

\begin{document}

\title{ST-LLM: Large Language Models Are Effective Temporal Learners
} 


\author{{Ruyang Liu}$^{1*\Diamond}$
    ~{Chen Li}$^{2*}$
    ~{Haoran Tang}$^{1}$
    ~{YiXiao Ge}$^{2}$
    ~{Ying Shan}$^{2}$ 
    ~{Ge Li \footnotesize{\Envelope}}$^{1}$\\ 
}

\authorrunning{Liu. et al.}
\institute{School of Electronic and Computer Engineering, Shenzhen Graduate School, Peking University
~~~ \email{\{ruyang@stu,hrtang@,geli@ece\}.pku.edu.cn}\\
\and
Applied Research Center (ARC), Tencent PCG
\email{\{palchenli,yixiaoge,yingsshan\}@tencent.com}}

\maketitle

\begin{abstract}
  Large Language Models (LLMs) have showcased impressive capabilities in text comprehension and generation, prompting research efforts towards video LLMs to facilitate human-AI interaction at the video level. However, how to effectively encode and understand videos in video-based dialogue systems remains to be solved. In this paper, we investigate a straightforward yet unexplored question: Can we feed all spatial-temporal tokens into the LLM, thus delegating the task of video sequence modeling to the LLMs? Surprisingly, this simple approach yields significant improvements in video understanding. Based upon this, we propose ST-LLM, an effective video-LLM baseline with \textbf{S}patial-\textbf{T}emporal sequence modeling inside \textbf{LLM}. Furthermore, to address the overhead and stability issues introduced by uncompressed video tokens within LLMs, we develop a dynamic masking strategy with tailor-made training objectives. For particularly long videos, we have also designed a global-local input module to balance efficiency and effectiveness. Consequently, we harness LLM for proficient spatial-temporal modeling, while upholding efficiency and stability. Extensive experimental results attest to the effectiveness of our method. Through a more concise model and training pipeline, ST-LLM establishes a new state-of-the-art result on VideoChatGPT-Bench and MVBench. Codes have been available at \href{https://github.com/TencentARC/ST-LLM}{https://github.com/TencentARC/ST-LLM}.
  
\end{abstract}

\section{Introduction}
\renewcommand{\thefootnote}{ } 
\footnotetext{* equal contribution~~~~$\Diamond$ Work done during internship at ARC Lab, Tencent PCG}

\label{sec:intro}
Large Language Models (LLMs), such as GPT \cite{brown2020language, achiam2023gpt}, PaLM \cite{anil2023palm, driess2023palm} and LLaMA \cite{touvron2023llama, touvron2023llama2}, have achieved remarkable success owing to their formidable language understanding and generation abilities, signaling a promising advancement towards artificial general intelligence. Driven by the widespread adoption of LLMs, research on Large Visual-Language Models (LVLMs) \cite{liu2023visual, zhu2023minigpt, dai2023instructblip} has emerged to extend the capabilities of LLMs to process visual signals, which have revealed proficiency in image-based conversations. However, in contrast to image inputs, videos feature heavier input and additional temporal information. Developing a large video-language model capable of effectively extracting meaningful spatial-temporal information from intricate video signals presents a significant challenge.

\begin{figure*}[t] 
    \centering
    \includegraphics[width=1\textwidth]{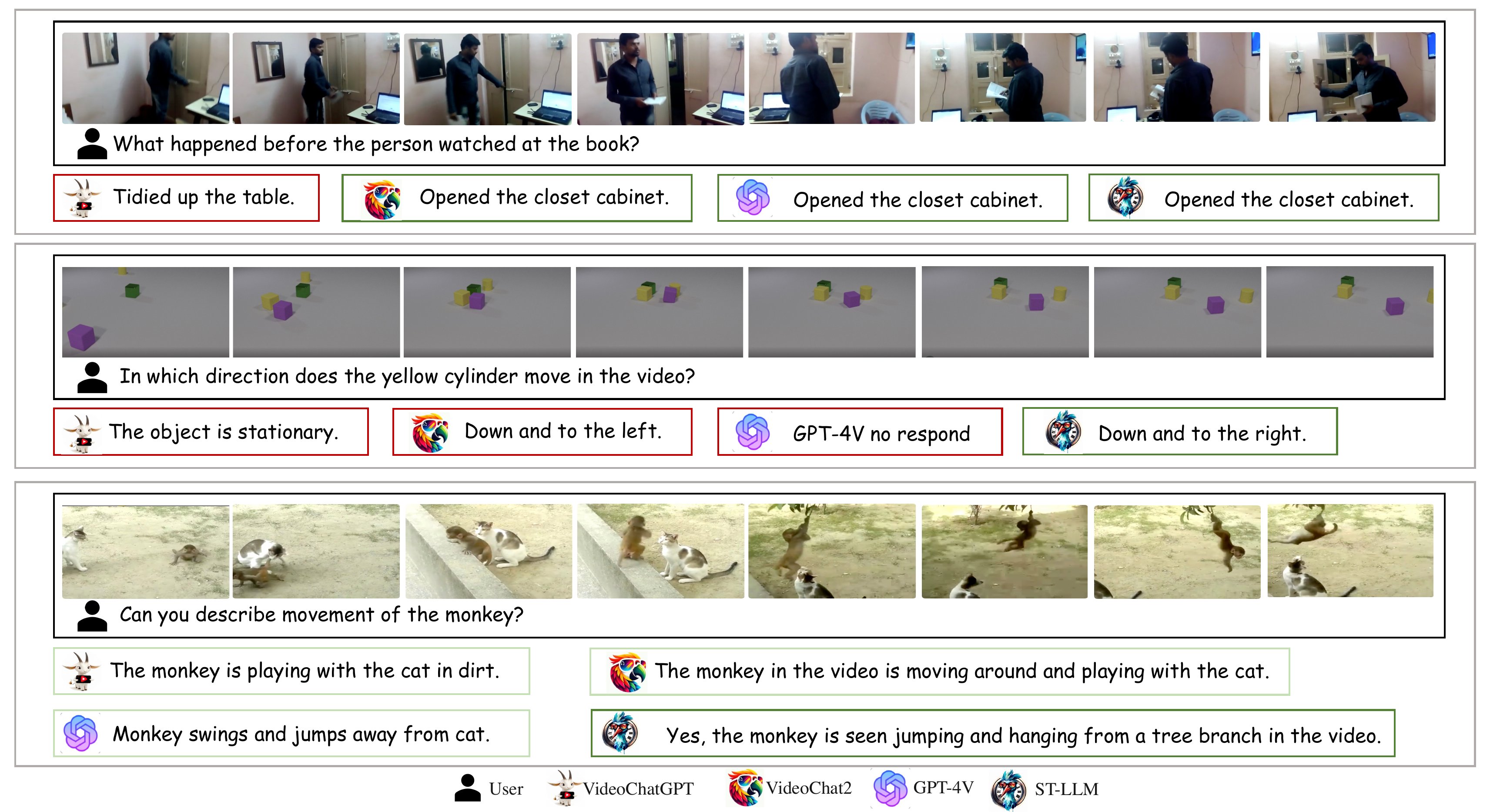} 
    \vspace{-2em}
    \caption{Qualitative comparisons among top-performance video LLMs. We illustrate two cases from MVBench \cite{li2023mvbench} (above) and one case from a YouTube video (below).}
    \label{image_intro1}
    \vspace{-1.3em}
  \end{figure*}

Recently, several preliminary attempts have surfaced aimed at extending LLMs to facilitate video conversation \cite{li2023videochat,luo2023valley,maaz2023video,zhang2023video,liu2023one,jin2023chat,huang2023vtimellm,li2023llama, li2023mvbench}, which unlock the capability of generating generic summaries of video content. Despite these efforts, existing video LLMs continue to exhibit shortcomings in achieving satisfactory performance in video comprehension, particularly in the realm of understanding content that relies heavily on temporal dynamics. As depicted in Fig. \ref{image_intro1}, contemporary top-performing video LLMs demonstrate robust understanding capabilities for actions reliant on static contexts. However, their proficiency in comprehending scenes involving motion remains constrained. They encounter challenges in discerning even the most basic direction of object movement, let alone the complex motion or scene transition.

\begin{figure*}[t] 
    \centering
    \includegraphics[width=1\textwidth]{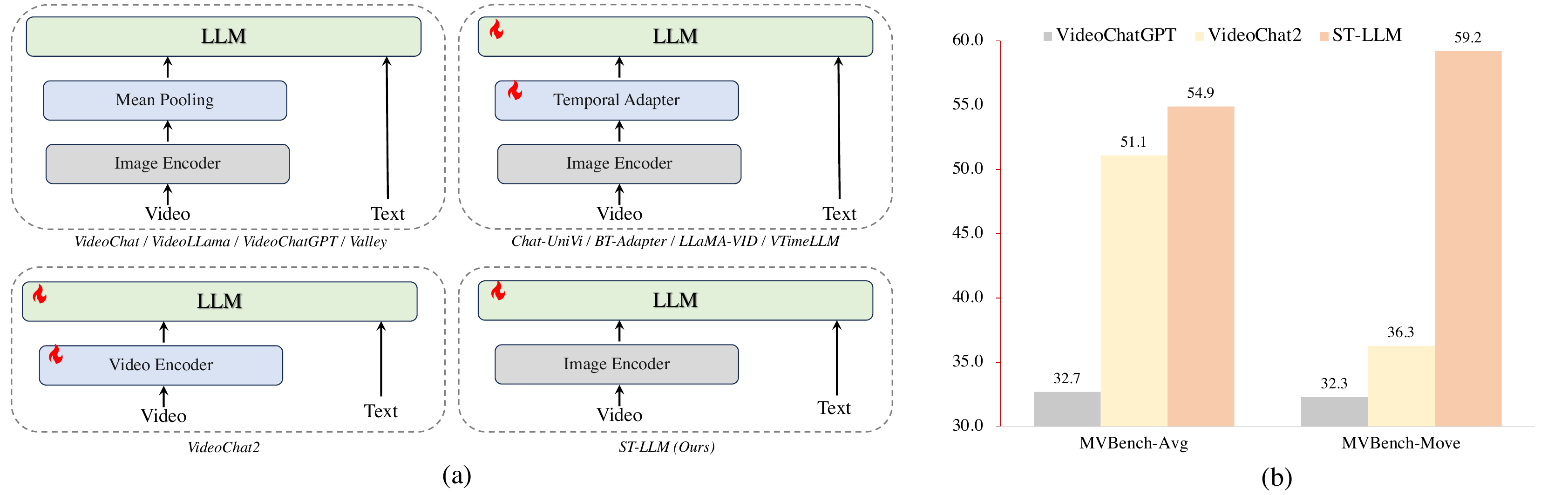} 
    \vspace{-2em}
    \caption{(a) Comparison among different video LLMs based on the way of inputting visual tokens. (b) Quantitative comparison of representative video LLMs, presenting the average results of all 20 metrics and temporal-sensitive motion-related metrics from MVBench, which encompass Moving Direction, Moving Count, and Moving Attribute.}
    \label{image_intro2}
    \vspace{-1.3em}
  \end{figure*}

This issue may stem from the inherent difficulty of temporal modeling within conversational systems. Historically, achieving robust video encoding has often required substantially more time and memory resources than images, which is impractical based on LLMs. As illustrated by Fig. \ref{image_intro2}(a), early video LLMs tend to adopt mean pooling on temporal dimension \cite{li2023videochat,luo2023valley,maaz2023video,zhang2023video}, a method that, while efficient, is inadequate for handling dynamic temporal sequences. 
Therefore, recent models frequently integrate additional structures for temporal sampling and modeling \cite{jin2023chat,huang2023vtimellm,li2023llama, li2023mvbench}. While these methods offer improved effectiveness compared to average pooling, they entail increased storage and demand extensive GPU time for training from scratch, often involving two or even three-stage pretraining to align new modules. 

Inspired by the recent success of joint spatial-temporal-text modeling in visual generation \cite{peebles2023scalable,sora24}, in this work, we investigate a simple yet unexplored idea: Given the robust sequence modeling capabilities inherent to LLMs, what if we input all visual tokens into the LLM and delegate the task of modeling spatial-temporal sequences to the LLM itself? Meanwhile, there are two additional challenges: (1) The inclusion of all visual tokens significantly increases the context length within LLMs, particularly for long videos, rendering the processing of a large number of frames unaffordable. (2) LLM may struggle to handle videos of varying lengths, potentially leading to hallucinations when there is a discrepancy between the number of frames in testing and training. 

To address these issues, we propose ST-LLM, a simple but powerful baseline of video LLM. As depicted in Fig. \ref{image_intro2}(a), with raw spatial-temporal tokens inside LLM, we harness the LLMs' robust sequence modeling capabilities for effective temporal modeling. Meanwhile, we introduce a novel dynamic video token masking strategy along with masking video modeling during training. With this approach, we reduce the length of sequences input to the LLM while significantly improving the robustness of videos of varying lengths during inference. To accommodate particularly long videos, we have devised a unique global-local input mechanism. This involves employing mean pooling of a large number of frames to generate residual input for a smaller subset of frames. Through this asymmetric design, we can process input from a large number of video frames while preserving the operations of modeling video tokens within LLMs.

As a result, we maintain input tokens of comparable length to most video LLMs while effectively harnessing the LLM's capacity to comprehend videos. Furthermore, as ST-LLM does not introduce additional modules, it does not necessitate expensive alignment pre-training. This enables ST-LLM to directly leverage existing state-of-the-art image conversational models, resulting in significantly reduced GPU time compared to other state-of-the-art video LLMs. Extensive experiments are conducted to verify the effectiveness of ST-LLM. Qualitatively, as depicted in Fig. \ref{image_intro1}, ST-LLM demonstrates a superior ability to understand dynamics compared to other video LLMs. Quantitatively, ST-LLM achieves new state-of-the-art performance across various contemporary video benchmarks, including MVBench \cite{li2023mvbench}, VideoChatGPT-Bench \cite{maaz2023video} and zero-shot QA-Evaluation. Specifically, as illustrated in Fig. \ref{image_intro2}(b), the superiority of ST-LLM is particularly evident in metrics related to the temporal-sensitive motion, which underscores the expertise of our model in temporal understanding.


The main contributions of our paper can be summarized as:
\vspace{-0.3em}
\begin{itemize}
        \item We propose ST-LLM, which, to our best knowledge, is the first open-source video LLM that explores spatial-temporal modeling within LLM.
        
        \item We present a dynamic video token masking strategy coupled with masked video modeling. Additionally, we introduce a global-local input mechanism for processing long videos. These innovations ensure the efficiency and robustness of spatial-temporal tokens within the LLM.

        \item Extensive experiments demonstrate the consistent superiority of ST-LLM over existing video LLMs across various video dialogue benchmarks, especially on tasks demanding robust temporal understanding.
\end{itemize}

\section{Related Works}
\noindent \textbf{LLMs and Image LLMs.}
Recent years have witnessed remarkable advancements in the evolution of Large Language Models (LLMs). Inspired by InstructGPT \cite{ouyang2022training} and the widely recognized commercial model ChatGPT~\cite{chatgpt}, the academic community has seen a proliferation of open-source LLMs, such as LLaMA \cite{touvron2023llama}, Alpaca \cite{taori2023stanford}, Vicuna \cite{chiang2023vicuna}, and LLaMA 2 \cite{touvron2023llama2}, which have become foundational components for a myriad of research endeavors. Fueled by abundant structured data and the escalation in model sizes, these LLMs have showcased remarkable capabilities in text comprehension and generation. Meanwhile, LLMs, being large-scale transformer models, inherently possess strong sequence modeling capabilities. This inspired us to explore the feasibility of entrusting the task of spatial-temporal sequence modeling to LLMs.
The success of LLMs has spurred increasing interest in the development of Multimodal Large Language Models (MLLMs). Notable breakthroughs include works such as Flamingo \cite{alayrac2022flamingo}, BLIP2 \cite{li2023blip}, and PaLM-E \cite{driess2023palm}, which have successfully bridged the gap between vision models and LLMs. Inspired by the concept of instruction tuning for LLMs, a series of works \cite{liu2023visual, zhu2023minigpt, dai2023instructblip, chen2023minigpt} leveraged open-source chatbots for visual instruction tuning, thus facilitating support for image-based conversations. It is noteworthy that in the pre-LLM era, video models were commonly constructed based on pre-trained 2D image models \cite{bertasius2021space, carreira2017quo, tran2015learning, liu2023revisiting}. This may be attributed to the significant computational demands of video processing, combined with the relatively limited scale and diversity of video data compared to images. Therefore, we believe that a well-pretrained image dialogue model could similarly offer training cost and performance benefits to video LLMs.

\vspace{0.3em}
  
\noindent \textbf{Video LLMs.}
The emergence of MLLMs quickly extended into the domain of video as well \cite{li2023videochat, maaz2023video, zhang2023video, luo2023valley, liu2023one, li2023mvbench, jin2023chat, li2023llama, huang2023vtimellm}. Early models like VideoChat \cite{li2023videochat}, VideoChatGPT \cite{maaz2023video} and Valley \cite{luo2023valley} generate video instruction tuning data through GPT to enable video conversations. At the model level, these approaches typically involve the use of mean pooling to aggregate the encoding results of individual frames before feeding them into the LLMs. While efficient, it is evident that this mean pooling method is inadequate for effective temporal modeling. Hence, some subsequent models have initiated exploration into adaptations of image models to video-specific requirements. For instance, BT-Adapter \cite{liu2023one} proposed a lightweight adapter to extend the capabilities of image LLMs. Chat-UniVi \cite{jin2023chat} introduces DPC-KNN to cluster dynamic visual tokens. Moreover, VideoChat2 \cite{li2023mvbench} deviated from using CLIP \cite{radford2021learning} and directly introduced a dedicated video encoder for video encoding. Although these methods offer improvements over mean pooling, the inclusion of additional modules typically necessitates intensive training. For example, in the case of VideoChat2, aligning the newly introduced video encoder entails a three-stage training involving 30 million visual-language samples. In contrast, ST-LLM adopts a more direct approach by entrusting the task of visual sequence modeling to LLMs, which is characterized by its conciseness, reduced training requirements, and, according to our experimental findings, superior effectiveness.

\noindent \textbf{Joint Spatial-Temporal Modeling.}
Prior to the emergence of LLMs, joint spatial-temporal modeling had demonstrated effectiveness in both video-only and video-language pretraining\cite{xue2022clip, tong2022videomae, li2023unmasked}. More recently, inspired by DIT \cite{peebles2023scalable}, the popular video generation model Sora \cite{sora24} has shown success in simultaneously processing video tokens and text using the transformer. Despite its potential, joint spatial-temporal-text modeling remains relatively uncommon in video LLMs. While VideoChat2 \cite{li2023mvbench} employed a joint-ST video encoder, it necessitated extensive pretraining. A closer relative, InstructBLIP \cite{dai2023instructblip} incorporated joint spatial-temporal sequences into LLM for zero-shot evaluation on video datasets. However, this approach led to significant hallucinations due to the absence of video instruction tuning. To the best of our knowledge, ST-LLM is the first open-source video LLM to adopt joint spatial-temporal-text modeling. Moreover, ST-LLM introduces a series of design innovations to mitigate the limitations associated with incorporating all video tokens within LLM.

\section{Methodology} \label{sec:method}
\begin{figure*}[t] 
    \centering
    \includegraphics[width=1\textwidth]{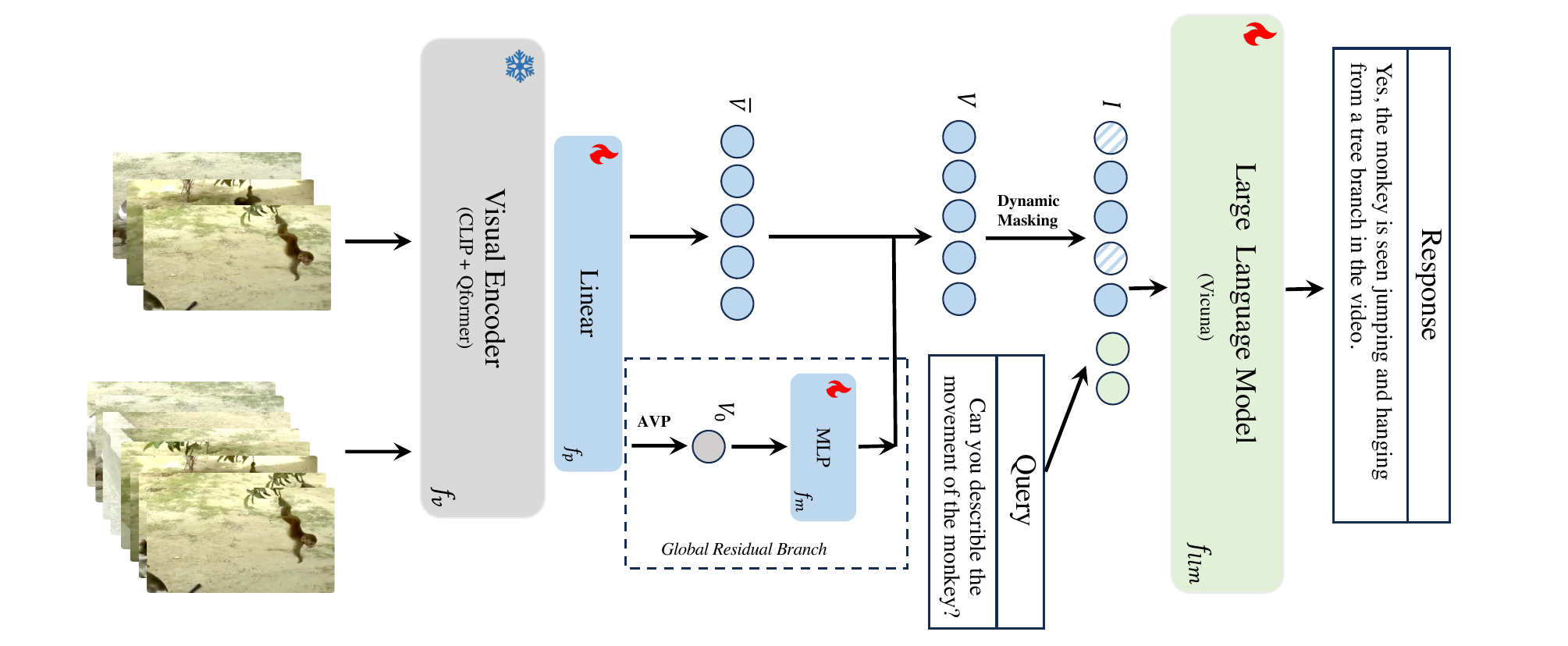} 
    \vspace{-2em}
    \caption{The overview of ST-LLM for generating responses on the input video and instructions. ST-LLM directly feeds the spatial-temporal sequence into the LLM and employs dynamic masking and global-local input for efficiency and robustness.}
    \label{image_method}
    \vspace{-1.3em}
  \end{figure*}

In this section, we elaborate on the structure and training methodology of ST-LLM. Firstly, we present a description of the model architecture, which incorporates all spatial-temporal tokens within the LLM, in Section \ref{sec:method1}. Subsequently, we delve into the masking mechanism and training objectives in Section \ref{sec:method2}. Finally, we introduce the global-local input approach in Section \ref{sec:method3}. The overarching framework of our method is illustrated in Fig. \ref{image_method}.

\subsection{Video Tokens Inside LLM} \label{sec:method1}
As depicted in Fig. \ref{image_method}, visually, ST-LLM closely resembles an image chatbot, consisting of a visual encoder, a linear projection, and an LLM. Instead of opting for the commonly used CLIP-L/14 \cite{radford2021learning}, we have selected BLIP-2 \cite{li2023blip} as our visual encoder, whose Q-Former efficiently compresses redundant image tokens into fewer visual tokens. To facilitate the input of raw videos for the image encoder, we adopt the strategy of treating each frame as an individual image. Given a video $v$ with $T$ frames, we assume that the image encoder $f_\mathrm{v}$ encodes each frame into $K$ tokens. Hence, we obtain all visual tokens $V$ to be fed into the LLM as follows:
\begin{equation}
    V =  f_\mathrm{p}(f_\mathrm{v}(v)),~ \operatorname{where}~ ~ V = \{V_i\}_{i=1}^T  , \ V_i = \{v_{(i,j)}\}_{j=1}^K,
\end{equation}
where $f_\mathrm{p}$ is the visual projection layer. Then, we concatenate all visual tokens to form a joint spatial-temporal \cite{bertasius2021space} visual sequence $V \in \mathbb{R}^{(T*K) \times D}$, where $D$ is the number of embedding dimensions. 
After being passed through the word embedding layers, the text tokens $C = \{c_{i}\}_{i=1}^N$ are directly concatenated with all spatial-temporal tokens, to organize the input $I$ to the LLM as follows:
\begin{equation}
    I = [V:C] = \{ v_{\scriptscriptstyle (1,1)}, v_{\scriptscriptstyle (1,2)}, ...,v_{\scriptscriptstyle (1,K)},v_{\scriptscriptstyle (2,1)}, ...,v_{\scriptscriptstyle (T,K)},c_1,c_2,...,c_N\}.
\end{equation}

Indeed, several alternative strategies exist for organizing the input tokens before feeding them into the LLM. These include: (1) Adding separator tokens between frame-wise tokens or visual-text tokens, to demarcate the boundaries between modalities or frames. (2) Adding spatial-temporal position embeddings specifically for visual tokens, to facilitate better spatial-temporal relationship understandings. However, our experiments indicate that simplicity often yields the best. Adding numerous separator tokens introduces additional overhead without necessarily improving performance. Moreover, LLMs come with Rotary Position Embeddings \cite{su2024roformer} that are already effective in distinguishing the positions of all spatial-temporal tokens and text tokens.

\subsection{Training with Dynamic Masking} \label{sec:method2}
With video tokens modeling inside LLM, we can leverage the potent sequence modeling capabilities inherent in LLMs to understand the information encapsulated within spatial-temporal sequences. However, integrating all video tokens into LLM significantly increases the context length, imposing a considerable computational burden and rendering it unfeasible to input additional frames for lengthy videos. Meanwhile, experimental findings suggest that during testing, utilizing uncompressed tokens is more sensitive to variations in the number of frames than mean pooling. Consequently, when a substantial dissonance exists between the number of frames in training and testing, a noticeable degradation in performance ensues.

To tackle the aforementioned issues, we first propose masking the visual tokens during training. Masking modeling has achieved significant success in natural language processing \cite{devlin2018bert} and video-language pretraining \cite{tong2022videomae, li2023unmasked, liu2023one}. Although masking is rarely employed in autoregressive LLMs, given our aim to leverage LLMs for encoding spatial-temporal sequences, we can still utilize masking modeling. Specifically, we keep the text tokens in $I$ unchanged, while applying a mask to the video tokens. This mask randomly masks $\rho$ of all video tokens, irrespective of their position across different frames. Different from previous works \cite{li2023unmasked, liu2023one}, we adopt a dynamic masking strategy, where the masking rate is randomly sampled from a normal distribution as follows:
\begin{equation}
    \rho \sim \mathbb{N}(0.5, \sigma), ~ ~ 0.3 \leq \rho \leq 0.7,
\end{equation}
where $\sigma$ represents the variance of the normal distribution. This strategy ensures that the length of the spatial-temporal sequence varies continuously while maintaining a consistently high average masking rate of 50\%. As a result, it minimizes training costs and notably enhances robustness during inference.

Furthermore, building upon dynamic masking, we have formulated the Masked Video Modeling (MVM) objective to encourage the LLM to grasp spatial-temporal dependencies. Unlike the expensive masked token recovery \cite{bao2021beit, peng2022beit}, we have opted for unmasked token reconstruction. Specifically, besides the masked sequence $I$, we conduct an extra non-grad forward pass for the unmasked sequence $\hat{I} = [\hat{V}:C]$, serving as the reference output. Then, we select unmasked tokens from $f_{llm}(I)$ and the corresponding tokens from $f_{llm}(\hat{I})$ based on their positions, computing the Mean Squared Error (MSE) between the selected pairs as follows:
\begin{equation}
    \mathcal{L}_{mvm}  = \frac{1}{(1-\rho) K T} \sum_{i=1}^{T} \sum_{j=1}^{K} (  v_{i,j}^{-1} - \hat{v}_{i,j}^{-1})^2, ~~(i,j) \notin M, 
 \end{equation}
where $M$ denotes the set of masked tokens' indices, $v_{i,j}^{-1}$ is from $f_{llm}(I)$, and $\mathcal{L}_{mvm}$ represents our MVM objective. Finally, our overall loss $\mathcal{L}$ is composed of  $\mathcal{L}_{mvm}$ and the LLM decoder loss $\mathcal{L}_{llm}$, yielding $\mathcal{L} = \mathcal{L}_{mvm}+\mathcal{L}_{llm}$. 
By integrating these two loss components, we can encourage the LLM to effectively respond to questions derived from the video content while improving its ability to model temporal and spatial dependencies.

\subsection{Global-Local Input} \label{sec:method3}
Although dynamic masking partially mitigates the challenge of processing lengthy input sequences, it fails to fully resolve the issue for extremely long videos that entail inputting numerous frames, rendering the context length still impractical to handle. Therefore, we have devised an additional module to tackle the issue of excessively long videos. To elaborate, given a lengthy video with a large $T$, we still commence by encoding each frame individually to derive $V$. Subsequently, we proceed to derive the global video representation $V_0$ through average pooling on the frame-wise tokens:
\begin{equation}
    v_{(0,j)} = \frac{1}{T} \sum_{i=1}^T v_{(i,j)},~~ V_0 = \{v_{(0,j)}\}_{j=1}^K.
 \end{equation}
Next, we average-sample $\overline{T}$ frames from the T total frames. All tokens from these $\overline{T}$ frames are concatenated to produce a joint spatial-temporal sequence $\overline{V} \in \mathbb{R}^{(\overline{T}*K) \times D}$, which serves as the local video representation. Finally, the global-local input for the LLM, denoted as $\overline{I}$, is constructed as follows:
\begin{equation}
    \overline{I} = [\overline{V}+f_{m}(V_0):C] ,
\end{equation}
where $f_{m}$ is a simple MLP projector with upsampling projection initialized with zeros. With this global-local input design, the low fps spatial-temporal sequences within the LLM can gradually incorporate information from the high fps branch. This approach allows the model to benefit from the LLM's ability to model temporal sequences within a limited context while also considering the global information of long videos.

\section{Experiments} \label{sec:experiment}
In this section, we have performed comprehensive experimental evaluations of ST-LLM, covering crucial settings, comparisons, and ablation studies. For a more detailed account of experimental settings, ablation studies, visualizations, and limitations analysis, please consult the appendix.

\vspace{-0.5em}
\subsection{Experiment Setup}
\noindent \textbf{Benchmarks.}
To evaluate the video understanding capabilities of ST-LLM, we primarily conduct assessments on three benchmarks: MVBench \cite{li2023mvbench}, VideoChatGPT Bench \cite{maaz2023video}, and zero-shot video QA benchmark.
MVBench comprises 20 challenging video tasks, each consisting of 200 samples in the form of multiple-choice questions. These tasks provide a comprehensive and objective assessment of a model's ability to understand videos. VideoChatGPT-Bench gathers videos from ActivityNet \cite{caba2015activitynet} and uses GPT to evaluate the quality of video conversations across five dimensions. Additionally, the zero-shot video QA benchmark entails GPT-based assessments of various open-source video QA datasets. 

\begin{table*}[t]
    \centering
    \setlength\tabcolsep{1pt}
    \caption{Comparisons on MVBench. Except for BLIP2 and Otter, all models are built upon LLaMA-1-7B for fair comparisons. "Avg" denotes the average of all 20 metrics. The leaderboards for each task can be found in the Appendix. The compared methods include: (1)Rd: random guesses; (2) mOI: mPLUG-Owl-I \cite{ye2023mplug}; (3) LMA: LLaMA-Adapter \cite{zhang2023llama} (4)B2: BLIP-2 \cite{li2023blip}; (5)OI: Otter-I \cite{li2023otter}; (6)MG: MiniGPT-4 \cite{zhu2023minigpt}; (7)IB: InstructBLIP \cite{dai2023instructblip}; (8)LV: LLaVA \cite{liu2023visual}; (9)OV: Otter-V \cite{li2023otter}; (10)mOV: mPLUG-Owl-V \cite{ye2023mplug}; (11)VCG: VideoChatGPT \cite{maaz2023video}; (12)VLM: VideoLLaMA \cite{zhang2023video}; (13)VC: VideoChat \cite{li2023videochat}; (14)VC2: VideoChat2 \cite{li2023mvbench}.}
    \vspace{-0.8em}
    \resizebox{1.0\textwidth}{!}{
        \begin{tabular}{l|c|c|c|c|c|c|c|c|c|c|c|c|c|c|c|c|c|c|c|c|c}
        \Xhline{1.0pt}
        \textbf{\tiny Metric} & Rd & mOI & LMA & B2 & OI & MG & IB & LV & OV & mOV & VCG & VLM & VC & VC2 & \cellcolor[RGB]{207,234,241}Ours\\
        \Xhline{1.0pt}
        AS & 25.0 & 25.0 & 23.0 & 24.5 & 34.5 & 16.0 & 20.0 & 28.0 & 23.0 & 22.0 & 23.5 & 27.5 & 33.5 & \textbf{66.0} & \cellcolor[RGB]{207,234,241}\textbf{66.0} \\
        AP & 25.0 & 20.0 & 28.0 & 29.0 & 32.0 & 18.0 & 16.5 & 39.5 & 23.0 & 28.0 & 26.0 & 25.5 & 26.5 & 47.5 & \cellcolor[RGB]{207,234,241}\textbf{53.5} \\
        AA & 33.3 & 44.5 & 51.0 & 33.5 & 39.5 & 26.0 & 46.0 & 63.0 & 27.5 & 34.0 & 62.0 & 51.0 & 50.0 & 83.5 & \cellcolor[RGB]{207,234,241}\textbf{84.0} \\
        FA & 25.0 & 27.0 & 30.0 & 17.0 & 30.5 & 21.5 & 24.5 & 30.5 & 27.0 & 29.0 & 22.5 & 29.0 & 33.5 & \textbf{49.5} & \cellcolor[RGB]{207,234,241}44.0 \\
        UA & 25.0 & 23.5 & 33.0 & 42.0 & 38.5 & 16.0 & 46.0 & 39.0 & 29.5 & 29.0 & 26.5 & 39.0 & 40.5 & \textbf{60.0} & \cellcolor[RGB]{207,234,241}58.5 \\
        OE & 33.3 & 36.0 & 53.5 & 51.5 & 48.5 & 29.5 & 51.0 & 53.0 & 53.0 & 40.5 & 54.0 & 48.0 & 53.0 & 58.0 & \cellcolor[RGB]{207,234,241}\textbf{80.5} \\
        OI & 25.0 & 24.0 & 32.5 & 26.0 & 44.0 & 25.5 & 26.0 & 41.0 & 28.0 & 27.0 & 28.0 & 40.5 & 40.5 & 71.5 & \cellcolor[RGB]{207,234,241}\textbf{73.5} \\
        OS & 33.3 & 34.0 & 33.5 & 31.0 & 29.5 & 13.0 & 37.5 & 41.5 & 33.0 & 31.5 & 40.0 & 38.0 & 30.0 & \textbf{42.5} & \cellcolor[RGB]{207,234,241}38.5 \\
        MD & 25.0 & 23.0 & 25.5 & 25.5 & 19.0 & 11.5 & 22.0 & 23.0 & 24.5 & 27.0 & 23.0 & 22.5 & 25.5 & 23.0 & \cellcolor[RGB]{207,234,241}\textbf{42.5} \\
        AL & 25.0 & 24.0 & 21.5 & 26.0 & 25.5 & 12.0 & 23.0 & 20.5 & 23.5 & 23.0 & 20.0 & 22.5 & 27.0 & 23.0 & \cellcolor[RGB]{207,234,241}\textbf{31.0} \\
        ST & 25.0 & 34.5 & 30.5 & 32.5 & 55.0 & 9.5 & 46.5 & 45.0 & 27.5 & 29.0 & 31.0 & 43.0 & 48.5 & \textbf{88.5} & \cellcolor[RGB]{207,234,241}86.5 \\
        AC & 33.3 & 34.5 & 29.0 & 25.5 & 20.0 & 32.5 & \textbf{42.5} & 34.0 & 26.0 & 31.5 & 30.5 & 34.0 & 35.0 & 39.0 & \cellcolor[RGB]{207,234,241}36.5 \\
        MC & 25.0 & 22.0 & 22.5 & 30.0 & 32.5 & 15.5 & 26.5 & 20.5 & 28.5 & 27.0 & 25.5 & 22.5 & 20.5 & 42.0 & \cellcolor[RGB]{207,234,241}\textbf{56.5} \\
        MA & 33.3 & 31.5 & 41.5 & 40.0 & 28.5 & 8.0 & 40.5 & 38.5 & 18.0 & 40.0 & 39.5 & 32.5 & 42.5 & 58.5 & \cellcolor[RGB]{207,234,241}\textbf{78.5} \\
        SC & 33.3 & 40.0 & 39.5 & 42.0 & 39.0 & 34.0 & 32.0 & 47.0 & 38.5 & 44.0 & \textbf{48.5} & 45.5 & 46.0 & 44.0 & \cellcolor[RGB]{207,234,241}43.0 \\
        FP & 25.0 & 24.0 & 25.0 & 27.0 & 28.0 & 26.0 & 25.5 & 25.0 & 22.0 & 24.0 & 29.0 & 32.5 & 26.5 & \textbf{49.0} & \cellcolor[RGB]{207,234,241}44.5 \\
        CO & 33.3 & 37.0 & 31.5 & 30.0 & 27.0 & 29.5 & 30.5 & 36.0 & 22.0 & 31.0 & 33.0 & 40.0 & 41.0 & 36.5 & \cellcolor[RGB]{207,234,241}\textbf{46.5} \\
        EN & 25.0 & 25.5 & 22.5 & 26.0 & 32.0 & 19.0 & 30.5 & 27.0 & 23.5 & 26.0 & 29.5 & 30.0 & 23.5 & \textbf{35.0} & \cellcolor[RGB]{207,234,241}34.5 \\
        ER & 20.0 & 21.0 & 28.0 & 37.0 & 36.5 & 9.9 & 30.5 & 26.5 & 19.5 & 20.5 & 26.0 & 21.0 & 23.5 & 40.5 & \cellcolor[RGB]{207,234,241}\textbf{41.5} \\
        CI & 30.9 & 37.0 & 32.0 & 31.0 & 36.5 & 3.0 & 38.0 & 42.0 & 19.0 & 29.5 & 35.5 & 37.0 & 36.0 & \textbf{65.5} & \cellcolor[RGB]{207,234,241}58.5 \\
        \rowcolor[RGB]{190,190,190} \textbf{Avg} & 27.3 & 29.4 & 31.7 & 31.4 & 33.5 & 18.8 & 32.5 & 36.0 & 26.8 & 29.7 & 32.7 & 34.1 & 35.5 & 51.1 & \cellcolor[RGB]{207,234,241}\textbf{54.9} \\
        \Xhline{1.0pt}
        \end{tabular}
    }
    
    \label{tab:mvpbench}
    \vspace{-0.5cm}
\end{table*}

\noindent \textbf{Implementation Details.}
In our study, we adopt state-of-the-art image dialogue models as the baseline and proceed directly with video instruction tuning. Unless otherwise specified, our model is initialized with InstructBLIP \cite{dai2023instructblip} in all experiments, whose LLM is Vicuna-v1.1 \cite{chiang2023vicuna} with 7B parameters. Additionally, we employ Minigpt4-v1 \cite{zhu2023minigpt} as a basic model to assess the effectiveness across various image models in Table~\ref{ablation:lora}. Following previous research \cite{lin2023video,jin2023chat,li2023llama}, we adopt full-finetuning on the LLM to ensure a fair comparison. The results of LoRA are also included as ablation results in Table~\ref{ablation:lora}. $\sigma$ is set as 0.1, ensuring that over 95\% of the mask rate falls within the range of 0.3 to 0.7.
During training, we sample 16 frames per video. Considering dynamic masking, an average of 256 tokens per video are input into the LLM, which is the same as LLaVA \cite{liu2023visual}. For all benchmarks during inference, frames are sampled at a frame rate of 1, and the number of local frames is kept between 4 and 16. This strategy closely aligns with practical scenarios and involves fewer frames than most video LLMs. For VideoChatGPT-Bench and the video QA benchmark, which may involve videos spanning several minutes, we utilize the global-local input. In this configuration, the global frame number is set to 64.
All videos are resized into 224*224. We train ST-LLM using lr of 2e-5 and batch size of 128 for 2 epochs.

\noindent \textbf{Training Datasets.}
Citing findings from Jin et al. \cite{jin2023chat}, the sequential process of first conducting image instruction tuning, followed by video instruction tuning and image-video joint pretraining, had a negligible impact on video dialogue performance. Hence, in light of InstructBLIP's prior training on extensive image data, we exclusively focus on video instruction tuning. Following \cite{li2023mvbench}, we leverage video instruction data sourced from a variety of datasets, including VideoChatGPT-100k \cite{maaz2023video}, VideoChat-11k \cite{li2023videochat}, Webvid \cite{bain2021frozen}, NExT-QA \cite{xiao2021next}, CLEVRER \cite{yi2019clevrer}, Kinetics-710 \cite{kay2017kinetics}, and Something-Something-2 \cite{goyal2017something}. All samples are uniformly formatted in accordance with \cite{li2023mvbench}. More details about the dataset and the organization of instructions can be found in the appendix. An experiment on joint image-video training is also presented in the appendix.

\begin{table*}[t]
\setlength{\tabcolsep}{3pt}
\centering
\caption{Comparisons on video-based generative performance benchmarking.}
\vspace{-0.8em}
\resizebox{0.95\textwidth}{!}{
\begin{tabular}{l|ccccc|c}
\toprule
  Method & Correct &  Detail & Context & Temporal & Consist & Mean Score \\ \midrule
 VideoLLaMA \cite{zhang2023video}    & 1.96 & 2.18 & 2.16 & 1.82  & 1.79 &  1.98 \\
 LLaMA-Adapter \cite{zhang2023llama} & 2.03 & 2.32 & 2.30 & 1.98  & 2.15 & 2.16 \\
 VideoChat \cite{li2023videochat}    & 2.23 & 2.50 & 2.53 & 1.94  & 2.24 & 2.29 \\
 VideoChatGPT \cite{maaz2023video}   & 2.40 & 2.52 & 2.62 & 1.98  & 2.37 & 2.38 \\
 BT-Adapter \cite{liu2023one}        & 2.68 & 2.69 & 3.27 & 2.34  & 2.46 & 2.69 \\
 VTimeLLM \cite{huang2023vtimellm}   & 2.78 & \textbf{3.10} & 3.40 & 2.49  & 2.47 & 2.85 \\
 Chat-UniVi \cite{jin2023chat}       & 2.89 & 2.91 & 3.46 & 2.89& \textbf{2.81} & 2.99 \\
 LLaMA-VID-7B \cite{li2023llama}     & 2.96 & 3.00 & 3.53 & 2.46  & 2.51 & 2.89 \\
 LLaMA-VID-13B \cite{li2023llama}    & 3.07 & 3.05 & 3.60 & 2.58  & 2.63 & 2.99 \\
 VideoChat2 \cite{li2023mvbench}     & 3.02 & 2.88 & 3.51 & 2.66  & \textbf{2.81} & 2.98 \\
 \rowcolor[RGB]{207,234,241} ST-LLM (Ours)  & \textbf{3.23} & 3.05 & \textbf{3.74} & \textbf{2.93}  & \textbf{2.81} & \textbf{3.15} \\

\bottomrule
\end{tabular}
}
\label{tab:vcgbench}
\vspace{-1.5em}
\end{table*}

\vspace{-0.5em}
\subsection{Quantitative Result}

\noindent \textbf{MVBench Performance.} 
The evaluation results on MVBench are presented in Table \ref{tab:mvpbench}. The evaluation metrics included Action Sequence (AS), Action Prediction (AP), Action Antonym (AA), Fine-grained Action (FA), Unexpected Action (UA), Object Existence (OE), Object Interaction (OI), Object Shuffle (OS), Moving Direction (MD), Action Localization (AL), Scene Transition (ST), Action Count (AC), Moving Count (MC), Moving Attribute (MA), State Change (SC), Fine-grained Pose (FP), Character Order (CO), Egocentric Navigation (EN), Episodic Reasoning (ER), Counterfactual Inference (CI), and the average of all 20 metrics (Avg). It is evident that on such a challenging benchmark, the performance of most multimodal LLMs falls significantly short of satisfactory levels, even barely surpassing random performance by less than 10\%. Our only formidable competitor is VideoChat2 \cite{li2023mvbench}. Through direct substitution of CLIP with a dedicated video encoder, coupled with a three-stage training process and comprehensive instruction tuning for the visual encoder, Q-Former, and LLM, VideoChat2 has yielded results that significantly surpass those of previous models. In contrast, we solely carried out single-stage video-only instruction tuning while maintaining the weights of CLIP and Q-Former frozen. Even under these conditions, ST-LLM still attained the top performance, with an average score surpassing VideoChat2 by a notable margin of 3.8\%. Specifically, ST-LLM showcased leading results across all 11 tasks, exhibiting a particularly significant advantage in motion-related tasks where other models struggled. In three motion-related tasks, ST-LLM achieved an average score of 59.2\%, far outperforming VideoChat2's 36.3\%. However, ST-LLM encounters challenges in fine-grained tasks, such as Fine-grained Action (FA) and Fine-grained Pose (FP). This could be attributed to CLIP's limitations as an image encoder, which may struggle to capture low-level spatial-temporal features required for these tasks.

\vspace{0.2em}

\begin{table*}[t]
\setlength{\tabcolsep}{3pt}
\centering
\caption{Comparisons on zero-shot question-answering, including MSVD-QA \cite{wu2017deep}, MSRVTT-QA \cite{xu2016msr}, and ActivityNet-QA \cite{caba2015activitynet}.}
\vspace{-0.8em}
\resizebox{0.88\textwidth}{!}{
\begin{tabular}{l|l|cc|cc|cc}
\toprule
\multirow{2}{*}{Method}  & \multirow{2}{*}{LLM}  & \multicolumn{2}{c|}{MSVD-QA} & \multicolumn{2}{c|}{MSRVTT-QA} & \multicolumn{2}{c}{ActivityNet-QA} \\
 & & Acc & Score & Acc & Score & Acc & Score \\ \hline
 FrozenBiLM \cite{yang2022zero} & DeBERTa-V2 & 32.2 & - & 16.8 & - & 24.7 & - \\ 
 VideoLLaMA \cite{zhang2023video} & Vicuna-7B & 51.6 & 2.5 & 29.6 & 1.8 & 12.4 & 1.1 \\ 
 LLaMA-Adapter \cite{zhang2023llama} & LLaMA-7B & 54.9 & 3.1 &43.8 &2.7 & 34.2 & 2.7 \\ 
 VideoChat \cite{li2023videochat} & Vicuna-7B & 56.3 & 2.8 & 45.0 & 2.5 & 26.5 & 2.2 \\ 
 VideoChatGPT \cite{maaz2023video} & Vicuna-7B & 64.9 & 3.3 & 49.3 & 2.8 & 35.2 & 2.7 \\ 
 BT-Adapter \cite{liu2023one} & Vicuna-7B & 67.5 & 3.7 & 57.0 & 3.2 & 45.7 & 3.2  \\
 Chat-UniVi \cite{jin2023chat} & Vicuna-7B & 65.0 & 3.6 & 54.6 & 3.1 & 45.8 & 3.2  \\
 LLaMA-VID \cite{li2023llama} & Vicuna-7B & 69.7 & 3.7 & 57.7 & 3.2 & 47.4 & 3.3  \\
 LLaMA-VID \cite{li2023llama} & Vicuna-13B & 70.0 & 3.7 & 58.9 & 3.3 & 47.5 & 3.3  \\
 VideoChat2 \cite{li2023mvbench} & Vicuna-7B & 70.0 & \textbf{3.9} & 54.1 & 3.3 & 49.1 & 3.3  \\
 \rowcolor[RGB]{207,234,241} ST-LLM (Ours) & Vicuna-7B & \textbf{74.6} & \textbf{3.9} & \textbf{63.2} & \textbf{3.4} & \textbf{50.9} & 3.3 \\ 
\bottomrule
\end{tabular}}
\label{tab:vcgqa}
\vspace{-1.5em}
\end{table*}

\noindent \textbf{VideoChatGPT-Bench Performance.} 
VideoChatGPT-Bench, also referred to as the Video-Based Generative Performance Benchmark, primarily evaluates the capability of video conversation. Utilizing GPT-3.5~\cite{chatgpt}, it assigns a score to the generated content across five aspects: Correctness of Information (Correct), Detail Orientation (Detail), Contextual Understanding (Context), Temporal Understanding (Temporal), and Consistency (Consist). As depicted in Table \ref{tab:vcgbench}, the recently proposed state-of-the-art models \cite{jin2023chat, li2023llama, li2023mvbench} exhibit similar performance levels on this benchmark. However, due to the inherent instability of GPT evaluation, there is considerable variation in the performance of each model across different tasks. Despite this variability, ST-LLM consistently performs admirably across all tasks, while demonstrating significant advantages in terms of the mean score compared to previous models.

\vspace{0.2em}

\noindent \textbf{Zero-shot Video-question Answering Performance.} In Table \ref{tab:vcgqa}, we present the zero-shot video question-answering performance on several commonly used video-text datasets, including MSVD-QA \cite{wu2017deep}, MSRVTT-QA \cite{xu2016msr}, and ActivityNet-QA \cite{caba2015activitynet}. Following the methodology outlined in \cite{maaz2023video}, we utilize GPT-3.5 to evaluate the accuracy and score of the generated results. As is depicted, 
STLLM demonstrates significant advantages on MSVD and MSRVTT, surpassing the previous SOTAs by 4.6\% and 4.3\% respectively. However, STLLM exhibits a slightly smaller advantage compared to VideoChat2 on the ActivityNet. This could be attributed to the requirement for a potent video encoder for this dataset.

\vspace{-0.5em}
\subsection{Ablations and Analysis}

\begin{figure}[ht]
  \begin{minipage}[b]{0.5\textwidth}
    \centering
    \captionof{table}{The ablation study on the methods of inputting video tokens to LLMs. The baseline configurations include no instruction tuning and mean pooling input.}
\resizebox{1.\textwidth}{!}{
\begin{tabular}{lc}
\toprule
 Method & Avg on MVBench \\ \hline
 \multicolumn{2}{l}{\textit{Instructblip Baseline}} \\
 Mean Pooling &  42.0 \\
 S-T Tokens in LLM & 35.5 \\ \hline
 \multicolumn{2}{l}{\textit{Video Instruction Tuning}} \\
 Mean Pooling &   47.8 \\
 S-T Tokens in LLM & 53.8 \\ 
 +Masking \& MVM Loss & \textbf{54.9} \\ 

\bottomrule
\end{tabular}}
\label{ablation:alltoken}
\vspace{-1em}
  \end{minipage}%
   \hfill
  \begin{minipage}[b]{0.47\textwidth}
\centering
    \includegraphics[width=1\textwidth]{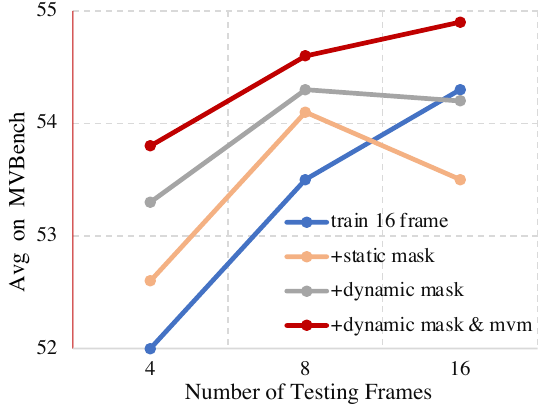} 
    \vspace{-2em}
    \caption{The ablation study on the effect of dynamic masking and MVM loss.}
    \label{ablation_mask}

\label{objective}
\vspace{-1em}
  \end{minipage}
\end{figure}

\begin{table*}[t]
\setlength{\tabcolsep}{3pt}
\centering
\caption{The ablation study on global-local input module. The ``Local'' and ``Global'' mean the joint temporal-temporal sequence and average pooling sequence respectively.}
\vspace{-0.8em}
\resizebox{1.0\textwidth}{!}{
\begin{tabular}{l|c|ccccc}
\toprule
 \multirow{2}{*}{Method} & \multirow{2}{*}{Avg on MVBench} & \multicolumn{5}{c}{Video-ChatGPT Bench} \\ 
 &  &  Correct & Detail & Context & Temporal & Consist \\ \hline
 Global Only &  48.3 & 3.01 & 2.92 & 3.58 & 2.61 & 2.61 \\
 Local Only &  \textbf{54.9} & 3.08 &  2.99 & 3.63 & 2.75 & 2.69 \\
 Local+Global (simply add) & 50.1  & 3.07 & 2.88 & 3.58 & 2.59 & 2.50 \\
 Local+Global (adapter) &  54.7 & \textbf{3.23} & \textbf{3.05} & \textbf{3.74} & \textbf{2.93}  & \textbf{2.81} \\
 
\bottomrule
\end{tabular}
}
\label{ablation:gl}
\vspace{-1.5em}
\end{table*}

\noindent \textbf{Ways of Inputting video tokens.} 
The core of this work lies in the endeavor to model joint spatial-temporal video tokens using LLM. Table \ref{ablation:alltoken} presents the results of different methods of inputting video tokens. As depicted, without video instruction tuning, the approach of joint spatial-temporal input is considerably less effective compared to directly averaging the input over the temporal dimension. This discrepancy might stem from inconsistencies between training and testing, resulting in severe hallucinations. However, after training, the effectiveness of the joint spatial-temporal input significantly surpasses that of the mean pooling input, indicating that LLMs can effectively model spatial-temporal sequences, and directly inputting all tokens is a superior approach compared to compression before input. Furthermore, incorporating dynamic masking and the MVM loss based on the joint spatial-temporal sequence has further improved the performance of our model.

\vspace{0.2em}

\noindent \textbf{Dynamic Masking Strategy and MVM Loss.}
In Fig. \ref{ablation_mask}, we conduct a more detailed ablation study on the effect of dynamic masking and MVM loss. It is evident that when the training frame count remains fixed, the performance of inputting all video tokens is not very robust. Particularly, when there is a discrepancy between the training and testing frame, the performance noticeably decreases. Meanwhile, when we randomly mask video tokens at a static 50\% rate, this issue persists, with effectiveness akin to training with a fixed 8-frame setup. However, employing dynamic masking shows some improvement in overall performance, and notably, the robustness to varying testing frame counts is significantly enhanced. Furthermore, the inclusion of the MVM loss built upon the masking further enhances performance, with greater improvements observed as the sequence length increases.

\vspace{0.2em}

\noindent \textbf{Global-Local Input Module.}
The ablation results on the global-local input module are presented in Table \ref{ablation:gl}, where we compare the performance of global features extracted from more frames and local features from fewer frames, as well as their combination. As shown, on MVBench, the performance of global features is significantly inferior to the joint S-T features, so combining global and local features did not lead to improvement. This may be attributed to the fact that MVBench mainly consists of short videos lasting a few to a dozen seconds, where a large number of frames might not be necessary. However, on VideoChatGPT Bench, where videos are parsed at the minute level, the discrepancy between global and local features has notably diminished, with their amalgamation resulting in a discernible enhancement. This underscores the efficacy of the Global-Local Input module in processing extended video content. Moreover, the direct incorporation of global and local features resulted in a performance decline, presumably due to its interference with the representation learning of the LLM.

\begin{table*}[t]
\setlength{\tabcolsep}{3pt}
\centering
\caption{The ablation study on the basic image models and the fine-tuning methodologies of LLM. We report the Avg on MVBench.}
\vspace{-0.8em}
\resizebox{1.0\textwidth}{!}{
\begin{tabular}{l|c|c|c|c}
\toprule
 Baseline Model & LLM & LLM Parameters & Meanpooling Baseline & ST-LLM Full Model \\ \hline
 \multirow{2}{*}{MiniGPT-4 \cite{zhu2023minigpt}}  & \multirow{2}{*}{Vicuna-7B v0} & LoRA & 40.9 & 45.6 \\
  &  & Full Fine-tuning & 46.4 & 50.5  \\ \hline
 \multirow{2}{*}{InstructBLIP \cite{dai2023instructblip}}  & \multirow{2}{*}{Vicuna-7B v1.1} & LoRA & 45.2 & 51.9  \\
  &  & Full Fine-tuning & 47.8 & 54.9  \\ 
\bottomrule
\end{tabular}
}
\label{ablation:lora}
\vspace{-1.5em}
\end{table*}

\begin{figure}[t]
  \begin{minipage}[b]{0.53\textwidth}
    \centering
    \captionof{table}{The analysis on the designs of inputting S-T video tokens into LLM.}
\resizebox{1.\textwidth}{!}{
\begin{tabular}{lc}
\toprule
 Method & Avg on MVBench  \\ \hline
 S-T Tokens in LLM (baseline) &  \textbf{54.9} \\ \hline
 +frame-frame separators &  54.0 \\
 +video-text separators & 54.8  \\
 +S-T position embeddings & 52.2 \\

\bottomrule
\end{tabular}}
\label{ablation:design1}
\vspace{-0.9em}
  \end{minipage}%
   \hfill
  \begin{minipage}[b]{0.45\textwidth}
    \centering
    \captionof{table}{The analysis on the designs of dynamic masking.}
\resizebox{1.\textwidth}{!}{
\begin{tabular}{lc}
\toprule
 Method & Avg on MVBench \\ \hline
 w/o masking (baseline) &  53.8 \\ \hline
 $\rho \sim \mathbb{U}(0.3, 0.7)$ & 54.2 \\
 $\rho \sim \mathbb{N}(0.5, 0.2)$ & 54.3  \\
 $\rho \sim \mathbb{N}(0.5, 0.1)$ & \textbf{54.9} \\
\bottomrule
\end{tabular}}
\label{ablation:design2}
\vspace{-0.9em}
  \end{minipage}

\end{figure}

\vspace{0.2em}

\noindent \textbf{Basic Image Models \& LLMs.}
In Table \ref{ablation:lora}, we conduct ablations on both the basic image chatbots and the LLM fine-tuning strategy. As depicted in the table, regardless of whether it's LoRA~\cite{hu2021lora} tuning or full fine-tuning, or different basic image models, ST-LLM demonstrates notable enhancements compared to the mean pooling baseline. This suggests that our approach can be adeptly applied across diverse scenarios, such as employing more resource-efficient LoRA tuning on larger models, or transferring to more potent image-dialogue models.

\vspace{0.2em}

\begin{figure*}[t] 
    \centering
    \includegraphics[width=1\textwidth]{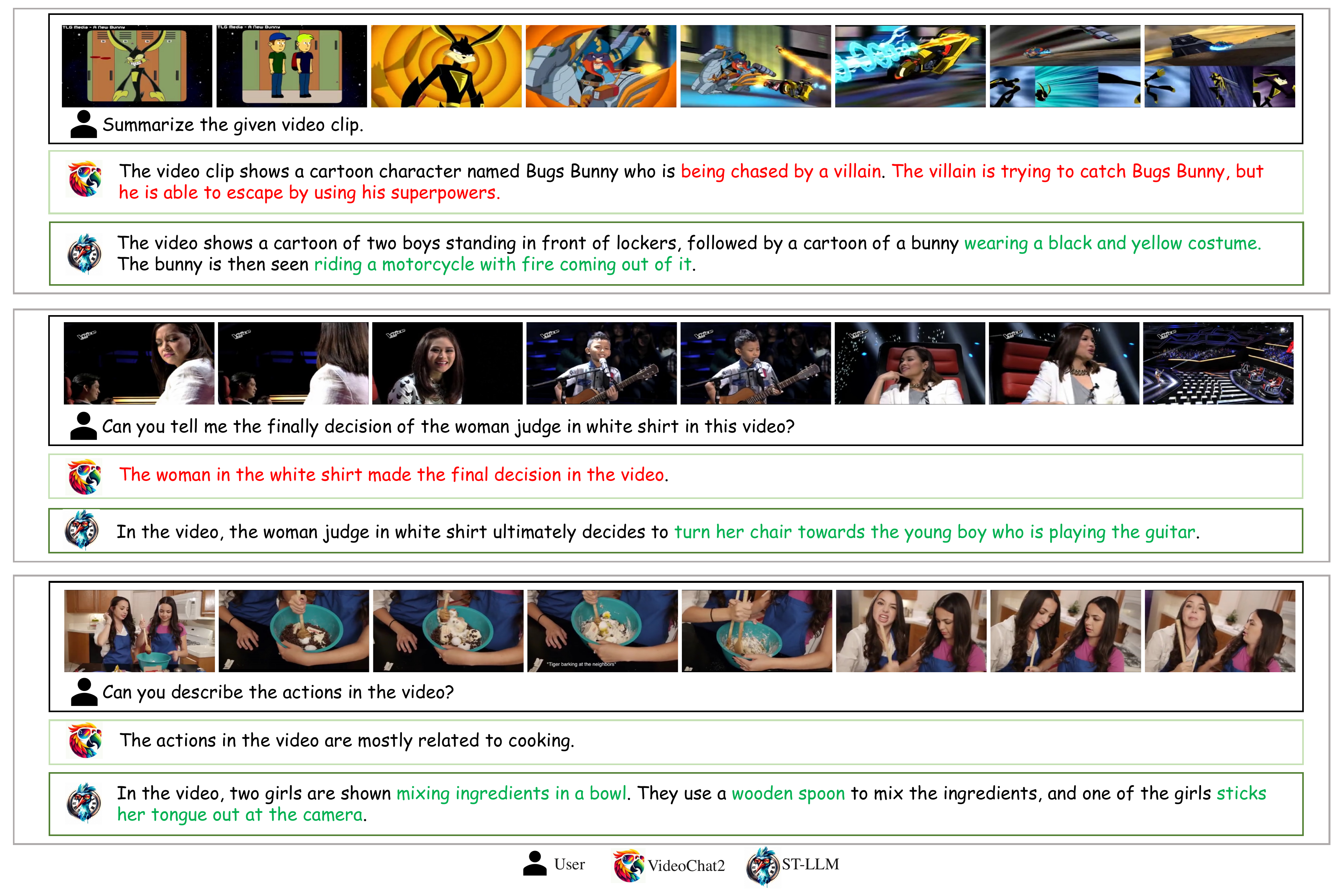} 
    \vspace{-2.3em}
    \caption{Qualitative results from real-world videos.}
    \label{visualize_maintext}
    \vspace{-1.3em}
  \end{figure*}
  
\noindent \textbf{Designs of ST-LLM.}
In Table \ref{ablation:design1} and Table \ref{ablation:design2}, we investigated several alternative design possibilities for ST-LLM. Primarily, we explored various separator designs. It's noteworthy that almost all cross-modal LLMs incorporate separator designs, with some models even integrating separators between multiple images or frames \cite{awadalla2023openflamingo, alayrac2022flamingo}. However, we discovered that when employing LLMs to model spatial-temporal sequences, these separators are superfluous. Subsequently, we address positional encodings. Numerous approaches integrate temporal positional encodings before inputting video tokens into LLMs \cite{luo2023valley,zhang2023video,maaz2023video}. However, in our methodology, we observed that incorporating absolute positional encodings to all visual tokens significantly impairs performance. Lastly, we consider the design of dynamic masking. Our findings suggest that all distributions of masking rates can yield some improvement, yet distributions with smaller standard deviations exhibit superior performance.

\subsection{Qualitative Results}
In Fig. \ref{visualize_maintext}, we present a qualitative comparison. As illustrated, ST-LLM excels in adhering to instructions and delivering precise responses. More importantly, ST-LLM showcases superior sensitivity to temporal sequences and actions, thereby validating the effectiveness of our modeling approach. For further quantitative results and failure cases, please refer to the appendix.

\section{Conclusion}
In this paper, we present ST-LLM, a straightforward yet robust video large language model. Our aim is to achieve effective video comprehension by leveraging LLM to model video tokens, marking the inception of a joint spatial-temporal-text modeling paradigm. Additionally, we introduce a dynamic masking strategy and a global-local input module to enhance this framework. Remarkably, ST-LLM excels in encoding and comprehending spatial-temporal sequences while addressing concerns related to efficiency, stability, and modeling lengthy videos, all with reduced training resource requirements. Through extensive experimental analysis, we demonstrate the effectiveness of ST-LLM and its internal design, resulting in state-of-the-art performance across multiple video LLM benchmarks.

%
\bibliographystyle{splncs04}
\bibliography{egbib}

\clearpage
\title{ST-LLM: Large Language Models Are Effective Temporal Learners \\ --------Supplementary Material--------}


\author{ }
\authorrunning{B. Ni et al.}
\institute{ }

\maketitle

This supplementary material contains additional details of the main paper, and provides more experiment analysis.

\section{Detailed Experiment Setup}\label{sec:detail}
Our basic image models comprise InstructBLIP \cite{dai2023instructblip} and MiniGPT4 \cite{zhu2023minigpt}, both of which are initialized by BLIP-2 \cite{li2023blip} and undergo additional image instruction tuning. BLIP-2 consists of a 39-layer EVA-CLIP \cite{sun2023eva} and a Q-former, compressing the 256 CLIP tokens into 32 visual tokens. In particular, InstructBlip's QFormer simultaneously accepts text input. In this study, we exclusively input the user's instruction into QFormer. Additionally, to enhance the video encoding capability, we experimented with inserting three layers of BT-Adapter \cite{liu2023one} alongside EVA-CLIP, while keeping the CLIP and Qformer frozen. The token sequence inputted into LLM follows the order: start token - visual token - text token - end token. No additional separators have been added. When employing the global-local input module, we initially feed all global tokens into the visual encoder, followed by sampling local frame-wise representations from the global representations. Training 2 epochs on 8 A100 GPUs with deepspeed zero-2 setting takes approximately 6 hours.

For different tasks, we employ distinct prompt designs. For MVBench, following Li \textit{et al.} \cite{li2023mvbench}, we have the question, system prompts (\textit{"Human: Carefully watch the video and pay attention to the cause and sequence of events, the detail and movement of objects, and the action and pose of persons. Based on your observations, select the best option that accurately addresses the question. Question:"}) and answer prompt (\textit{"Only give the best option. Assistant: Best option:(''}). For GPT-based evaluation, we have reorganized the instructions as question, system prompt (\textit{"Carefully watch the video and pay attention to the cause and sequence of events, the detail and movement of objects, and the action and pose of persons. Based on your observations, give your answer that best addresses the question. Human:"}) and answer prompt (\textit{"Only give the best option. Assistant: ''}). The input text for LLM is formed as system prompt - question - answer prompt. However, different from \cite{li2023mvbench}, only the question is fed into the former without including the system prompt. This approach aims to maximize the extraction of visually associated tokens with the keywords.

\begin{table*}[t]
\setlength{\tabcolsep}{3pt}
\centering
\caption{The ablation study on the image-video joint pretraining. The video data is the default dataset mentioned earlier, while the image data is from LLaVA-150k. We report the average score on MVBench.}
\vspace{-0.8em}
\resizebox{1.0\textwidth}{!}{
\begin{tabular}{l|c|c|c|c}
\toprule
 Baseline Model & Pretrain Data & Image-Only & Video-Only & Joint Image-Video  \\ \hline
 MiniGPT-4 \cite{zhu2023minigpt}  & MiniGPT4-3k & 39.4 & 50.5 & 51.9 \\ \hline
 \multirow{2}{*}{InstructBLIP \cite{dai2023instructblip}}  & LLaVA-150k and & \multirow{2}{*}{42.1} & \multirow{2}{*}{54.9} & \multirow{2}{*}{53.1}  \\
 & 10 image-text datasets & & & \\
\bottomrule
\end{tabular}
}
\label{ablation:joint}
\end{table*}

\begin{figure}[ht]
  \begin{minipage}[b]{0.53\textwidth}
  \centering
    \includegraphics[width=1\textwidth]{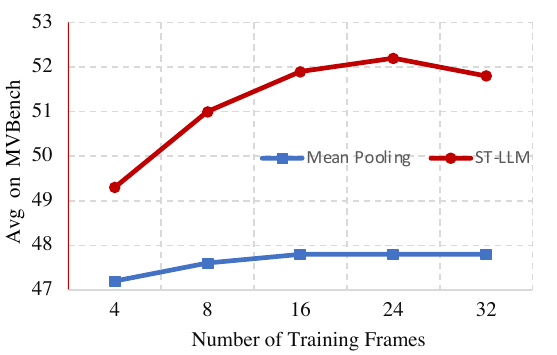 } 
    \vspace{-2em}
    \caption{The ablation study on the effect of the number of training frames.}
    \label{ablation_noframe}
  \end{minipage}%
   \hfill
  \begin{minipage}[b]{0.43\textwidth}
\centering
    \captionof{table}{The ablation study on the testing strategy. ``Fixed'' means adopting the same fixed inference frames as training. We report the average score on MVBench.}
\resizebox{1.\textwidth}{!}{
\begin{tabular}{l|c|c}
\toprule
 Method & Fixed & FPS1 \\ \hline
 Mean Pooling & 47.8 & 47.5 \\ \hline
 Train 8 Frames & 52.8 & 52.5 \\ 
 +Masking \& MVM Loss & 53.1 & 53.5 \\ \hline
 Train 16 Frames & 54.3 & 53.9 \\ 
 +Masking \& MVM Loss & 54.8 & 54.9 \\ 
\bottomrule
\end{tabular}}
\label{ablation:fps1}

  \end{minipage}
\end{figure}

\section{Additional Ablations}\label{sec:aabaltion}
\noindent \textbf{Image-Video Joint Pretraining.}
Many video LLMs utilize both image and video data during alignment pre-training or instruction tuning \cite{li2023mvbench,li2023llama, jin2023chat, lin2023video}, with some models even regarding this as a core contribution \cite{jin2023chat, lin2023video}. Moreover, image data has been demonstrated to be effective for video tasks in downstream tasks \cite{bain2021frozen, li2023unmasked}. Therefore, we examined the results of image-video joint training on ST-LLM in Table \ref{ablation:joint}. As shown, Image-video joint training leads to improvements in MiniGPT4 but causes a decrease in performance for InstructBlip. This discrepancy arises because InstructBlip has already been trained on rich image data, and repeating the training may lead to catastrophic forgetting. Conversely, MiniGPT4 has limited training data, making the incorporation of image data effective. This suggests that image-video joint training can be effective for our model. However, if a well-pretrained image base model is already utilized, including image training may be unnecessary.

\vspace{0.2em}

\noindent \textbf{Scale Up Input Frames.}
In Fig. \ref{ablation_noframe}, we explore the performance of ST-LLM with varying numbers of input frames. To facilitate large-scale training, we present the results of Lora for ST-LLM. As depicted, for mean pooling input, the results show little sensitivity to increasing input frame counts. However, with ST-LLM, a noticeable improvement is observed, particularly when the frame count is low. This highlights ST-LLM's capability to effectively scale up frame counts. Nevertheless, when the frame count becomes very high, further scaling does not yield additional improvements. This may be attributed to MVBench primarily comprising short videos that do not necessitate numerous frames, or it could indicate the current challenges LLMs face in modeling such extended visual contexts solely through simple instruction tuning.

\vspace{0.2em}

\noindent \textbf{Inference Strategy.}
In this paper, we employed a testing strategy with a frame rate of 1 while constraining the upper and lower bounds of the frame count. This approach, which captures different frame counts for videos of varying lengths, is more realistic and challenging. As illustrated in Table \ref{ablation:fps1}, generally, training and testing consistency is crucial, and fixed frame counts yield better results than a frame rate of 1. However, with our comprehensive method, the results with a frame rate of 1 are marginally superior to those with fixed frame counts.

\section{MVBench Leaderboards}\label{sec:leaderborads}
To provide a clearer comparison of different tasks on MVBench, we present the leaderboards of various tasks in Table \ref{tab:leaderboard}. Overall, ST-LLM achieves the highest rank across 11 tasks. In tasks related to actions (a)(b)(c)(d)(j), ST-LLM generally performs well and outperforms VideoChat2 overall. In tasks related to movement (i)(m)(p), ST-LLM's performance is particularly outstanding, with significant advantages in each task. The exceptional performance in both action and movement tasks underscores ST-LLM's superior temporal modeling capability. In tasks related to objects (f)(g)(h), ST-LLM also showcases significant advantages, which could be attributed to the wealth of knowledge inherited from image LLMs. However, ST-LLM exhibits a notable performance gap compared to VideoChat2 on fine-grained tasks (d)(o). This discrepancy may arise from the traditional video encoders being more effective than CLIP in low-level spatiotemporal modeling. Additionally, both ST-LLM and VideoChat2 display poor performance on certain tasks (l)(n), which could be attributed to catastrophic forgetting induced during training.

\section{Additional Qualitative Results}\label{sec:visual_appendix}
\begin{figure*}[!htb] 
    \centering
    \includegraphics[width=1\textwidth]{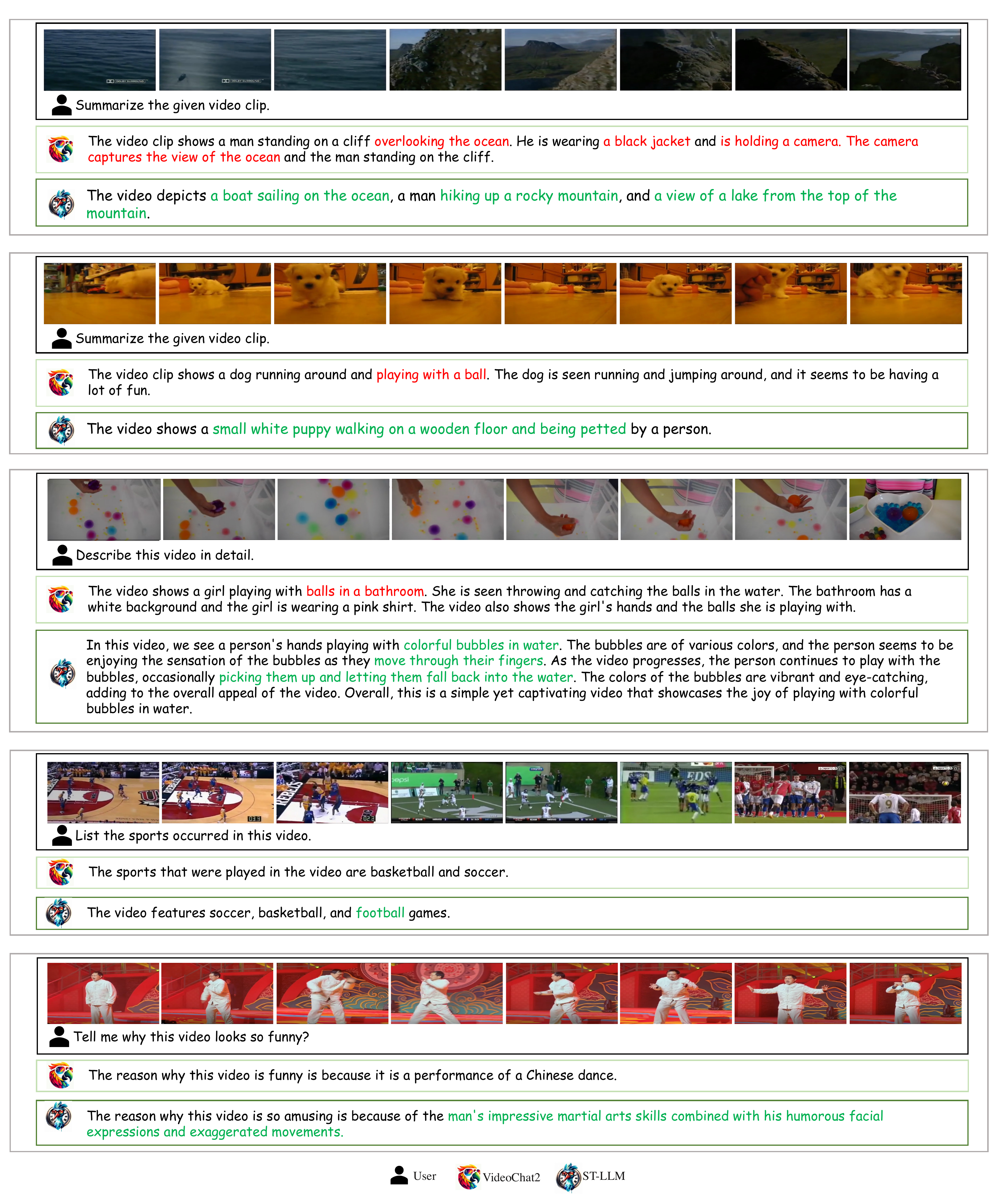} 
    \vspace{-1.3em}
    \caption{More qualitative results from real-world videos.}
    \label{visualize_appendix1}
  \end{figure*}

\begin{figure*}[t] 
    \centering
    \includegraphics[width=1\textwidth]{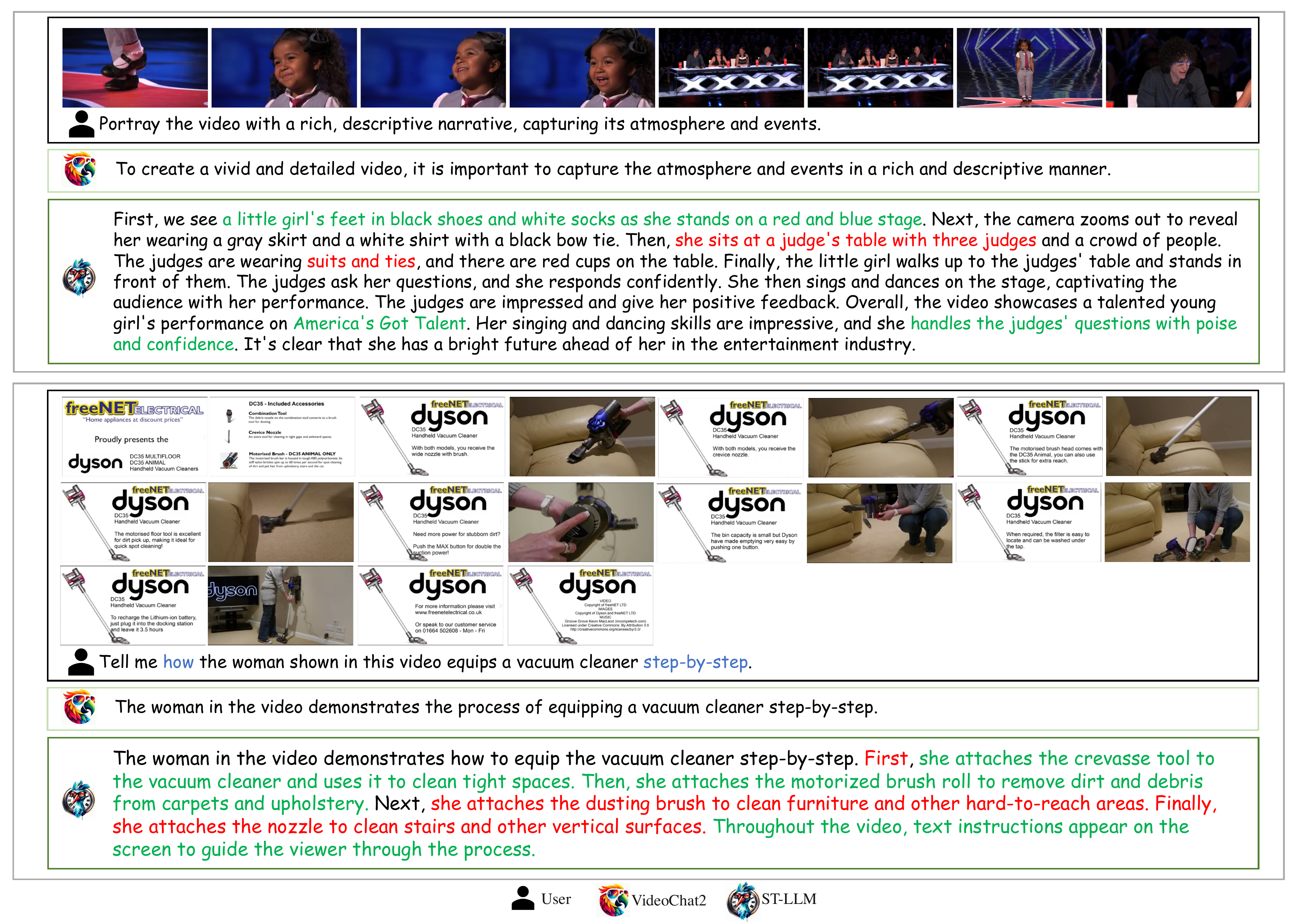} 
    \vspace{-1.3em}
    \caption{Qualitative results of detailed description or step-by-step interpretation for long videos. }
    \label{visualize_appendix2}
    \vspace{-0.5em}
  \end{figure*}

In Fig. \ref{visualize_appendix1} and Fig. \ref{visualize_appendix2}, we present additional visualization results, encompassing video content summaries, detailed video descriptions, inference of video content, and fine-grained descriptions of long videos. Across these aspects, ST-LLM demonstrates exceptional performance. Notably, ST-LLM excels in adhering strictly to user instructions while accurately describing video content and capturing dynamic information. In particular, for detailed and step-by-step video descriptions, VideoChat2 falls short, whereas ST-LLM outperforms it by a considerable margin.

\begin{table*}[tp]
\centering
\begin{minipage}[t]{0.236\textwidth}
    \vspace{0pt}
    \centering
    \setlength\tabcolsep{4.0pt}
    \resizebox{1\linewidth}{!}{
        \begin{tabular}{c|l|c}
        \textbf{Rank} & \multicolumn{1}{c|}{\textbf{Model}} & \textbf{Acc} \\
        \Xhline{1.0pt}
        \rowcolor{blue!20}\textbf{1} &  \textbf{\textcolor{red}{ST-LLM}} & \textbf{66.0}\\
        \rowcolor{blue!12}\textbf{2} &  \textbf{VideoChat2} & \textbf{66.0}\\
        \rowcolor{blue!6}\textbf{3} &  \textbf{Otter-I} & \textbf{34.5}\\
        4 &  VideoChat & 33.5\\
        5 &  LLaVA & 28.0 \\
        6 &  VideoLLaMA & 27.5 \\
        7 &  mPLUG-Owl-I & 25.0 \\
        8 &  BLIP2 & 24.5 \\
        9 &  VideoChatGPT & 23.5 \\
        10 &  LLaMA-Adapter & 23.0 \\
        11 &  InstructBLIP & 20.0 \\
        \end{tabular}
    }
    \subcaption{{\tiny Action Sequence}}
\end{minipage}
\hfill
\begin{minipage}[t]{0.236\textwidth}
    \vspace{0pt}
    \centering
    \setlength\tabcolsep{4.0pt}
    \resizebox{1\linewidth}{!}{
        \begin{tabular}{c|l|c}
        \textbf{Rank} & \multicolumn{1}{c|}{\textbf{Model}} & \textbf{Acc} \\
        \Xhline{1.0pt}
        \rowcolor{blue!20}\textbf{1} &  \textbf{\textcolor{red}{ST-LLM}} & \textbf{53.5}\\
        \rowcolor{blue!12}\textbf{2} &  \textbf{VideoChat2} & 47.5 \\
        \rowcolor{blue!6}\textbf{3} &  \textbf{LLaVA} & \textbf{39.5}\\
        4 &  Otter-I & 32.0\\
        5 &  BLIP2 & 29.0 \\
        6 &  LLaMA-Adapter & 28.0 \\
        7 &  VideoChat & 26.5 \\
        8 &  VideoChatGPT & 26.0 \\
        9 &  VideoLLaMA & 25.5 \\
        10 &  mPLUG-Owl-I & 20.0 \\
        11 &  MiniGPT-4 & 18.0 \\
        \end{tabular}
    }
    \subcaption{{\tiny Action Prediction}}
\end{minipage}
\hfill
\begin{minipage}[t]{0.236\textwidth}
    \vspace{0pt}
    \centering
    \setlength\tabcolsep{4.0pt}
    \resizebox{1\linewidth}{!}{
        \begin{tabular}{c|l|c}
        \textbf{Rank} & \multicolumn{1}{c|}{\textbf{Model}} & \textbf{Acc} \\
        \Xhline{1.0pt}
        \rowcolor{blue!20}\textbf{1} &  \textbf{\textcolor{red}{ST-LLM}} & \textbf{84.0}\\
        \rowcolor{blue!12}\textbf{2} &  \textbf{VideoChat2} & \textbf{83.5}\\
        \rowcolor{blue!6}\textbf{3} &  \textbf{LLaVA} & \textbf{63.0}\\
        4 &  VideoChatGPT & 62.0\\
        5 &  VideoChat & 56.0 \\
        6 &  LLaMA-Adapter & 51.0 \\
        7 &  VideoLLaMA & 51.0 \\
        8 &  InstructBLIP & 46.0 \\
        9 &  mPLUG-Owl-I & 44.5 \\
        10 &  Otter-I & 39.5 \\
        11 &  BLIP2 & 33.5 \\
        \end{tabular}
    }
    \subcaption{{\tiny Action Antonym}}
\end{minipage}
\hfill
\begin{minipage}[t]{0.236\textwidth}
    \vspace{0pt}
    \centering
    \setlength\tabcolsep{4.0pt}
    \resizebox{1\linewidth}{!}{
        \begin{tabular}{c|l|c}
        \textbf{Rank} & \multicolumn{1}{c|}{\textbf{Model}} & \textbf{Acc} \\
        \Xhline{1.0pt}
        \rowcolor{blue!20}\textbf{1} &  \textbf{VideoChat2} & \textbf{49.5}\\
        \rowcolor{blue!12}\textbf{2} &  \textbf{\textcolor{red}{ST-LLM}} & \textbf{44.0}\\
        \rowcolor{blue!6}\textbf{3} &  \textbf{VideoChat} & \textbf{33.5}\\
        4 &  Otter-I & 30.5 \\
        5 &  LLaVA & 30.5 \\
        6 &  LLaMA-Adapter & 30.0 \\
        7 &  VideoLLaMA & 29.0 \\
        8 &  mPLUG-Owl-I & 27.0 \\
        9 &  InstructBLIP & 24.5 \\
        10 &  VideoChatGPT & 22.5 \\
        11 &  MiniGPT-4 & 21.5 \\
        \end{tabular}
    }
    \subcaption{{\tiny Fine-grained Action}}
\end{minipage}

\begin{minipage}[t]{0.236\textwidth}
    \vspace{0pt}
    \centering
    \setlength\tabcolsep{4.0pt}
    \resizebox{1\linewidth}{!}{
        \begin{tabular}{c|l|c}
        \textbf{Rank} & \multicolumn{1}{c|}{\textbf{Model}} & \textbf{Acc} \\
        \Xhline{1.0pt}
        \rowcolor{blue!20}\textbf{1} &  \textbf{VideoChat2} & \textbf{60.0}\\
        \rowcolor{blue!12}\textbf{2} &  \textbf{\textcolor{red}{ST-LLM}} & \textbf{58.5}\\
	\rowcolor{blue!6}\textbf{3} &  \textbf{InstructBLIP} & \textbf{46.0}\\
	4 &  BLIP2 & 42.0\\
	5 &  VideoChat & 40.5 \\
	6 &  LLaVA & 39.0 \\
	7 &  VideoLLaMA & 39.0 \\
	8 &  Otter-I & 38.5 \\
	9 &  LLaMA-Adapter & 33.0 \\
	10 &  VideoChatGPT & 26.5 \\
	11 &  mPLUG-Owl-I & 23.5 \\
        \end{tabular}
    }
    \subcaption{{\tiny Unexpected Action}}
\end{minipage}
\hfill
\begin{minipage}[t]{0.236\textwidth}
    \vspace{0pt}
    \centering
    \setlength\tabcolsep{3.pt}
    \resizebox{1\linewidth}{!}{
        \begin{tabular}{c|l|c}
        \textbf{Rank} & \multicolumn{1}{c|}{\textbf{Model}} & \textbf{Acc} \\
        \Xhline{1.0pt}
        \rowcolor{blue!20}\textbf{1} &  \textbf{\textcolor{red}{ST-LLM}} & \textbf{80.5}\\
        \rowcolor{blue!12}\textbf{2} &  \textbf{VideoChat2} & \textbf{58.0}\\
	\rowcolor{blue!6}\textbf{3} &  \textbf{VideoChatGPT} & \textbf{54.0}\\
	4 &  LLaMA-Adapter & 53.5\\
	5 &  LLaVA & 53.0 \\
	6 &  VideoChat & 53.0 \\
	7 &  BLIP2 & 51.5 \\
	8 &  InstructBLIP & 51.0 \\
	9 &  Otter-I & 48.5 \\
	10 &  VideoLLaMA & 48.0 \\
	11 &  mPLUG-Owl-I & 36.0 \\
        \end{tabular}
    }
    \subcaption{{\tiny Object Existence}}
\end{minipage}
\hfill
\begin{minipage}[t]{0.236\textwidth}
    \vspace{0pt}
    \centering
    \setlength\tabcolsep{4.0pt}
    \resizebox{1\linewidth}{!}{
        \begin{tabular}{c|l|c}
        \textbf{Rank} & \multicolumn{1}{c|}{\textbf{Model}} & \textbf{Acc} \\
        \Xhline{1.0pt}
        \rowcolor{blue!20}\textbf{1} &  \textbf{\textcolor{red}{ST-LLM}} & \textbf{73.5}\\
        \rowcolor{blue!12}\textbf{2} &  \textbf{VideoChat2} & \textbf{71.5}\\
	\rowcolor{blue!6}\textbf{3} &  \textbf{Otter-I} & \textbf{44.0}\\
	4 &  LLaVA & 41.0\\
	5 &  VideoLLaMA & 40.5 \\
	6 &  VideoChat & 40.5 \\
	7 &  LLaMA-Adapter & 32.5 \\
	8 &  VideoChatGPT & 28.0 \\
	9 &  BLIP2 & 26.0 \\
	10 &  InstructBLIP & 26.0 \\
	11 &  MiniGPT-4 & 25.5 \\
        \end{tabular}
    }
    \subcaption{{\tiny Object Interaction}}
\end{minipage}
\hfill
\begin{minipage}[t]{0.236\textwidth}
    \vspace{0pt}
    \centering
    \setlength\tabcolsep{4.0pt}
    \resizebox{1\linewidth}{!}{
        \begin{tabular}{c|l|c}
        \textbf{Rank} & \multicolumn{1}{c|}{\textbf{Model}} & \textbf{Acc} \\
        \Xhline{1.0pt}
        \rowcolor{blue!20}\textbf{1} &  \textbf{VideoChat2} & \textbf{42.5}\\
	\rowcolor{blue!12}\textbf{2} &  \textbf{LLaVA} & \textbf{41.5}\\
	\rowcolor{blue!6}\textbf{3} &  \textbf{VideoChatGPT} & \textbf{40.0}\\
        4 &  \textcolor{red}{ST-LLM} & 38.5 \\
	5 &  VideoLLaMA & 38.0 \\
	6 &  InstructBLIP & 37.5 \\
	7 &  mPLUG-Owl-I & 34.0 \\
	8 &  LLaMA-Adapter & 33.5 \\
	9 &  BLIP2 & 31.0 \\
	10 &  VideoChat & 30.0 \\
	11 &  Otter-I & 29.5 \\

        \end{tabular}
    }
    \subcaption{\tiny Object Shuffle}
\end{minipage}

\begin{minipage}[t]{0.236\textwidth}
    \vspace{0pt}
    \centering
    \setlength\tabcolsep{3.0pt}
    \resizebox{1\linewidth}{!}{
        \begin{tabular}{c|l|c}
        \textbf{Rank} & \multicolumn{1}{c|}{\textbf{Model}} & \textbf{Acc} \\
        \Xhline{1.0pt}
        \rowcolor{blue!20}\textbf{1} &  \textbf{\textcolor{red}{ST-LLM}} & \textbf{42.5}\\
        \rowcolor{blue!12}\textbf{2} &  \textbf{LLaMA-Adapter} & \textbf{25.5}\\
	\rowcolor{blue!6}\textbf{3} &  \textbf{BLIP2} & \textbf{25.5}\\
	4 &  VideoChat & 25.5\\
	5 &  VideoChat2 & 23.0 \\
	6 &  VideoChatGPT & 23.0 \\
	7 &  mPLUG-Owl-I & 23.0 \\
	8 &  LLaVA & 23.0 \\
	9 &  VideoLLaMA & 22.5 \\
	10 &  InstructBLIP & 22.0 \\
	11 &  Otter-I & 19.0 \\
        \end{tabular}
    }
    \subcaption{{\tiny Moving Direction}}
\end{minipage}
\hfill
\begin{minipage}[t]{0.236\textwidth}
    \vspace{0pt}
    \centering
    \setlength\tabcolsep{4.0pt}
    \resizebox{1\linewidth}{!}{
        \begin{tabular}{c|l|c}
        \textbf{Rank} & \multicolumn{1}{c|}{\textbf{Model}} & \textbf{Acc} \\
        \Xhline{1.0pt}
        \rowcolor{blue!20}\textbf{1} &  \textbf{\textcolor{red}{ST-LLM}} & \textbf{31.0}\\
        \rowcolor{blue!12}\textbf{2} &  \textbf{VideoChat} & \textbf{27.0}\\
	\rowcolor{blue!6}\textbf{3} &  \textbf{BLIP2} & \textbf{26.0}\\
	4 &  Otter-I & 25.5\\
	5 &  mPLUG-Owl-I & 24.0 \\
	6 &  VideoChat2 & 23.0 \\
	7 &  InstructBLIP & 23.0 \\
	8 &  VideoLLaMA & 22.5 \\
	9 &  LLaMA-Adapter & 21.5 \\
	10 &  LLaVA & 20.5 \\
	11 &  VideoChatGPT & 20.0 \\
        \end{tabular}
    }
    \subcaption{{\tiny Action Localization}}
\end{minipage}
\hfill
\begin{minipage}[t]{0.236\textwidth}
    \vspace{0pt}
    \centering
    \setlength\tabcolsep{4.0pt}
    \resizebox{1\linewidth}{!}{
        \begin{tabular}{c|l|c}
        \textbf{Rank} & \multicolumn{1}{c|}{\textbf{Model}} & \textbf{Acc} \\
        \Xhline{1.0pt}
        \rowcolor{blue!20}\textbf{1} &  \textbf{VideoChat2} & \textbf{88.5}\\
        \rowcolor{blue!12}\textbf{2} &  \textbf{\textcolor{red}{ST-LLM}} & \textbf{86.5}\\
	\rowcolor{blue!6}\textbf{3} &  \textbf{Otter-I} & \textbf{55.0}\\
	4 &  VideoChat & 48.5\\
	5 &  InstructBLIP & 46.5 \\
	6 &  LLaVA & 45.0 \\
	7 &  VideoLLaMA & 43.0 \\
	8 &  mPLUG-Owl-I & 34.5 \\
	9 &  BLIP2 & 32.5 \\
	10 &  VideoChatGPT & 31.0 \\
	11 &  LLaMA-Adapter & 30.5 \\
        \end{tabular}
    }
    \subcaption{{\tiny Scene transition}}
\end{minipage}
\hfill
\begin{minipage}[t]{0.236\textwidth}
    \vspace{0pt}
    \centering
    \setlength\tabcolsep{4.0pt}
    \resizebox{1\linewidth}{!}{
        \begin{tabular}{c|l|c}
        \textbf{Rank} & \multicolumn{1}{c|}{\textbf{Model}} & \textbf{Acc} \\
        \Xhline{1.0pt}
        \rowcolor{blue!20}\textbf{1} &  \textbf{InstructBLIP} & \textbf{42.5}\\
	\rowcolor{blue!12}\textbf{2} &  \textbf{VideoChat2} & \textbf{39.0}\\
	\rowcolor{blue!6}\textbf{3} &  \textbf{\textcolor{red}{ST-LLM}} & \textbf{36.5}\\
        4 &  VideoChat & 35.0\\
	5 &  mPLUG-Owl-I & 34.5 \\
	6 &  LLaVA & 34.0 \\
	7 &  VideoLLaMA & 34.0 \\
	8 &  MiniGPT-4 & 32.5 \\
	9 &  VideoChatGPT & 30.5 \\
	10 &  LLaMA-Adapter & 29.0 \\
	11 &  BLIP2 & 25.5 \\
        \end{tabular}
    }
    \subcaption{{\tiny Action Count}}
\end{minipage}

\begin{minipage}[t]{0.236\textwidth}
    \vspace{0pt}
    \centering
    \setlength\tabcolsep{4.0pt}
    \resizebox{1\linewidth}{!}{
        \begin{tabular}{c|l|c}
        \textbf{Rank} & \multicolumn{1}{c|}{\textbf{Model}} & \textbf{Acc} \\
        \Xhline{1.0pt}
        \rowcolor{blue!20}\textbf{1} &  \textbf{\textcolor{red}{ST-LLM}} & \textbf{56.5}\\
        \rowcolor{blue!12}\textbf{2} &  \textbf{VideoChat2} & \textbf{42.0}\\
	\rowcolor{blue!6}\textbf{3} &  \textbf{Otter-I} & \textbf{32.5}\\
	4 &  BLIP2 & 30.0\\
	5 &  InstructBLIP & 26.5 \\
	6 &  VideoChatGPT & 25.5 \\
	7 &  VideoLLaMA & 22.5 \\
	8 &  LLaMA-Adapter & 22.5 \\
	9 &  mPLUG-Owl-I & 22.0 \\
	10 &  LLaVA & 20.5 \\
	11 &  VideoChat & 20.5 \\
        \end{tabular}
    }
    \subcaption{\tiny Moving Count}
\end{minipage}
\hfill
\begin{minipage}[t]{0.236\textwidth}
    \vspace{0pt}
    \centering
    \setlength\tabcolsep{4.0pt}
    \resizebox{1\linewidth}{!}{
        \begin{tabular}{c|l|c}
        \textbf{Rank} & \multicolumn{1}{c|}{\textbf{Model}} & \textbf{Acc} \\
        \Xhline{1.0pt}
        \rowcolor{blue!20}\textbf{1} &  \textbf{VideoChatGPT} & \textbf{48.5}\\
	\rowcolor{blue!12}\textbf{2} &  \textbf{LLaVA} & \textbf{47.0}\\
	\rowcolor{blue!6}\textbf{3} &  \textbf{VideoChat} & \textbf{46.0}\\
	4 &  VideoLLaMA & 45.5 \\
	5 &  VideoChat2 & 44.0 \\
        6 &  \textcolor{red}{ST-LLM} & 43.0 \\
	7 &  BLIP2 & 42.0 \\
	8 &  mPLUG-Owl-I & 40.0 \\
	9 &  LLaMA-Adapter & 39.5 \\
	10 &  Otter-I & 39.0 \\
	11 &  MiniGPT-4 & 34.0 \\
        \end{tabular}
    }
    \subcaption{\tiny State Change}
\end{minipage}
\hfill
\begin{minipage}[t]{0.236\textwidth}
    \vspace{0pt}
    \centering
    \setlength\tabcolsep{4.0pt}
    \resizebox{1\linewidth}{!}{
        \begin{tabular}{c|l|c}
        \textbf{Rank} & \multicolumn{1}{c|}{\textbf{Model}} & \textbf{Acc} \\
        \Xhline{1.0pt}
        \rowcolor{blue!20}\textbf{1} &  \textbf{VideoChat2} & \textbf{49.0}\\
        \rowcolor{blue!12}\textbf{2} &  \textbf{\textcolor{red}{ST-LLM}} & \textbf{44.5}\\
	\rowcolor{blue!6}\textbf{3} &  \textbf{VideoLLaMA} & \textbf{32.5}\\
	4 &  VideoChatGPT & 29.0\\
	5 &  Otter-I & 28.0 \\
	6 &  BLIP2 & 27.0 \\
	7 &  VideoChat & 26.5 \\
	8 &  MiniGPT-4 & 26.0 \\
	9 &  InstructBLIP & 25.5 \\
	10 &  LLaMA-Adapter & 25.0 \\
	11 &  LLaVA & 25.0 \\
        \end{tabular}
    }
    \subcaption{{\tiny Fine-grained Pose}}
\end{minipage}
\hfill
\begin{minipage}[t]{0.236\textwidth}
    \vspace{0pt}
    \centering
    \setlength\tabcolsep{3.0pt}
    \resizebox{1\linewidth}{!}{
        \begin{tabular}{c|l|c}
        \textbf{Rank} & \multicolumn{1}{c|}{\textbf{Model}} & \textbf{Acc} \\
        \Xhline{1.0pt}
        \rowcolor{blue!20}\textbf{1} &  \textbf{\textcolor{red}{ST-LLM}} & \textbf{78.5}\\
        \rowcolor{blue!12}\textbf{2} &  \textbf{VideoChat2} & \textbf{58.5}\\
	\rowcolor{blue!6}\textbf{3} &  \textbf{VideoChat} & \textbf{42.5}\\
	4 &  LLaMA-Adapter & 41.5\\
	5 &  InstructBLIP & 40.5 \\
	6 &  BLIP2 & 40.0 \\
	7 &  VideoChatGPT & 39.5 \\
	8 &  LLaVA & 38.5 \\
	9 &  VideoLLaMA & 32.5 \\
	10 &  mPLUG-Owl-I & 31.5 \\
	11 &  Otter-I & 28.5 \\
        \end{tabular}
    }
    \subcaption{\tiny Moving Attribute}
\end{minipage}

\begin{minipage}[t]{0.236\textwidth}
    \vspace{0pt}
    \centering
    \setlength\tabcolsep{4.0pt}
    \resizebox{1\linewidth}{!}{
        \begin{tabular}{c|l|c}
        \textbf{Rank} & \multicolumn{1}{c|}{\textbf{Model}} & \textbf{Acc} \\
        \Xhline{1.0pt}
        \rowcolor{blue!20}\textbf{1} &  \textbf{\textcolor{red}{ST-LLM}} & \textbf{46.5}\\
        \rowcolor{blue!12}\textbf{2} &  \textbf{VideoChat} & \textbf{41.0}\\
	\rowcolor{blue!6}\textbf{3} &  \textbf{VideoLLaMA} & \textbf{40.0}\\
	4 &  mPLUG-Owl-I & 37.0\\
	5 &  VideoChat2 & 36.5 \\
	6 &  LLaVA & 36.0 \\
	7 &  VideoChatGPT & 33.0 \\
	8 &  LLaMA-Adapter & 31.5 \\
	9 &  BLIP2 & 30.0 \\
	10 &  InstructBLIP & 30.0 \\
	11 &  MiniGPT-4 & 29.5 \\
        \end{tabular}
    }
    \subcaption{{\tiny Character Order}}
\end{minipage}
\hfill
\begin{minipage}[t]{0.236\textwidth}
    \vspace{0pt}
    \centering
    \setlength\tabcolsep{4.0pt}
    \resizebox{1\linewidth}{!}{
        \begin{tabular}{c|l|c}
        \textbf{Rank} & \multicolumn{1}{c|}{\textbf{Model}} & \textbf{Acc} \\
        \Xhline{1.0pt}
        \rowcolor{blue!20}\textbf{1} &  \textbf{VideoChat2} & \textbf{35.0}\\
        \rowcolor{blue!12}\textbf{2} &  \textbf{\textcolor{red}{ST-LLM}} & \textbf{34.5}\\
	\rowcolor{blue!6}\textbf{3} &  \textbf{Otter-I} & \textbf{32.0}\\
	4 &  VideoLLaMA & 30.0\\
	5 &  VideoChatGPT & 29.5 \\
	6 &  LLaVA & 27.0 \\
	7 &  BLIP2 & 26.0 \\
	8 &  mPLUG-Owl-I & 25.5 \\
	9 &  InstructBLIP & 25.5 \\
	10 &  VideoChat & 23.5 \\
	11 &  LLaMA-Adapter & 22.5 \\
        \end{tabular}
    }
    \subcaption{\tiny Egocentric Navigation}
\end{minipage}
\hfill
\begin{minipage}[t]{0.236\textwidth}
    \vspace{0pt}
    \centering
    \setlength\tabcolsep{4.0pt}
    \resizebox{1\linewidth}{!}{
        \begin{tabular}{c|l|c}
        \textbf{Rank} & \multicolumn{1}{c|}{\textbf{Model}} & \textbf{Acc} \\
        \Xhline{1.0pt}
        \rowcolor{blue!20}\textbf{1} &  \textbf{\textcolor{red}{ST-LLM}} & \textbf{41.5}\\
        \rowcolor{blue!12}\textbf{2} &  \textbf{VideoChat2} & \textbf{40.5}\\
	\rowcolor{blue!6}\textbf{3} &  \textbf{BLIP2} & \textbf{37.0}\\
	4 &  InstructBLIP & 30.5\\
	5 &  Otter-I & 29.0 \\
	6 &  LLaMA-Adapter & 28.0 \\
	7 &  LLaVA & 26.5 \\
	8 &  VideoChatGPT & 26.0 \\
	9 &  VideoChat & 23.5 \\
	10 &  mPLUG-Owl-I & 21.0 \\
	11 &  VideoLLaMA & 21.0 \\
        \end{tabular}
    }
    \subcaption{{\tiny Episodic Reasoning}}
\end{minipage}
\hfill
\begin{minipage}[t]{0.236\textwidth}
    \vspace{0pt}
    \centering
    \setlength\tabcolsep{4.0pt}
    \resizebox{1\linewidth}{!}{
        \begin{tabular}{c|l|c}
        \textbf{Rank} & \multicolumn{1}{c|}{\textbf{Model}} & \textbf{Acc} \\
        \Xhline{1.0pt}
        \rowcolor{blue!20}\textbf{1} &  \textbf{VideoChat2} & \textbf{65.5}\\
        \rowcolor{blue!12}\textbf{2} &  \textbf{\textcolor{red}{ST-LLM}} & \textbf{58.5}\\
	\rowcolor{blue!6}\textbf{3} &  \textbf{LLaVA} & \textbf{42.0}\\
	4 &  InstructBLIP & 38.0\\
	5 &  mPLUG-Owl-I & 37.0 \\
	6 &  VideoLLaMA & 37.0 \\
	7 &  Otter-I & 36.5 \\
	8 &  VideoChat & 36.0 \\
	9 &  VideoChatGPT & 35.5 \\
	10 &  LLaMA-Adapter & 32.0 \\
	11 &  BLIP2 & 31.0 \\
        \end{tabular}
    }
    \subcaption{{\tiny Counterfactual Inference}}
\end{minipage}
\vspace{-0.1cm}
\caption{Leaderboards of different tasks in MVBench.}
\vspace{-0.3cm}
\label{tab:leaderboard}
\end{table*}

\section{Limitations and Future Work}\label{sec:limit_appendix}
From the quantitative experiments conducted earlier, it is evident that ST-LLM does not excel at fine-grained tasks. This suggests that without a robust foundation in low-level spatiotemporal modeling, LLMs also struggle with particularly fine-grained spatiotemporal modeling. Moreover, from Fig \ref{visualize_appendix2}, it is apparent that in the detailed description or step-by-step interpretation of long videos, although ST-LLM outperforms VideoChat2, it still falls short of being satisfactory, and some hallucinations may still occur. This indicates that there is still much progress to be made in comprehensive
 video understanding. Additionally, for exceptionally long videos, such as those at the level of movies, ST-LLM still lacks particularly effective strategies to maintain both performance and low computational cost.

 In this paper, we have demonstrated the feasibility and high effectiveness of inputting all video tokens into LLM and utilizing LLM for spatiotemporal modeling. In future work, we will continue to delve deeper into this direction from three perspectives. Firstly, we will explore better methods to leverage LLM for spatiotemporal modeling, such as enhancing training objectives. Secondly, we will focus on enhancing efficiency and improving adaptability for long videos, such as compressing spatiotemporal sequences and selecting key tokens. Lastly, we aim to scale up our model by attempting to use larger models, more data, and more extensive pre-training.
 
\end{document}